\documentclass[11pt,letterpaper]{mystyle}
\usepackage{fancyhdr}

\usepackage{tikz}
\usepackage[utf8]{inputenc} 
\usepackage[T1]{fontenc}
\usepackage[numbers]{natbib}
\usepackage{adjustbox}
\usepackage{breakurl}    
\usepackage{hyperref}
\usepackage[edges]{forest}
\usepackage{subcaption}
\usepackage{soul}
\usepackage{multirow}
\usepackage[utf8]{inputenc} 
\usepackage{booktabs}       
\usepackage{amsfonts}       
\usepackage{nicefrac}      
\usepackage{microtype}
\usepackage{parskip}
\usepackage{xcolor}        
\usepackage{colortbl}
\usepackage{enumitem}
\usepackage[numbers]{natbib}
\usepackage{amssymb}
\usepackage{wrapfig}
\usepackage{courier}
\usepackage{bxcoloremoji}
\usepackage{CJKutf8}
\usepackage{ragged2e}
\usepackage{makecell}
\usepackage{longtable}
\usepackage[useregional]{datetime2}
\usepackage{threeparttable}
\usepackage{xspace}
\definecolor{tamuMaroon}{HTML}{500000}
\colorlet{tamuGrayMaroon}{tamuMaroon!15!black}
\newcommand{\github}{\raisebox{-1.5pt}{\includegraphics[height=1.05em]{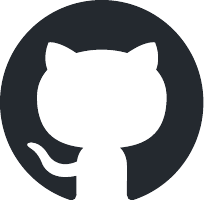}}}
\newcommand{\huggingface}{\raisebox{-1.5pt}{\includegraphics[height=1.05em]{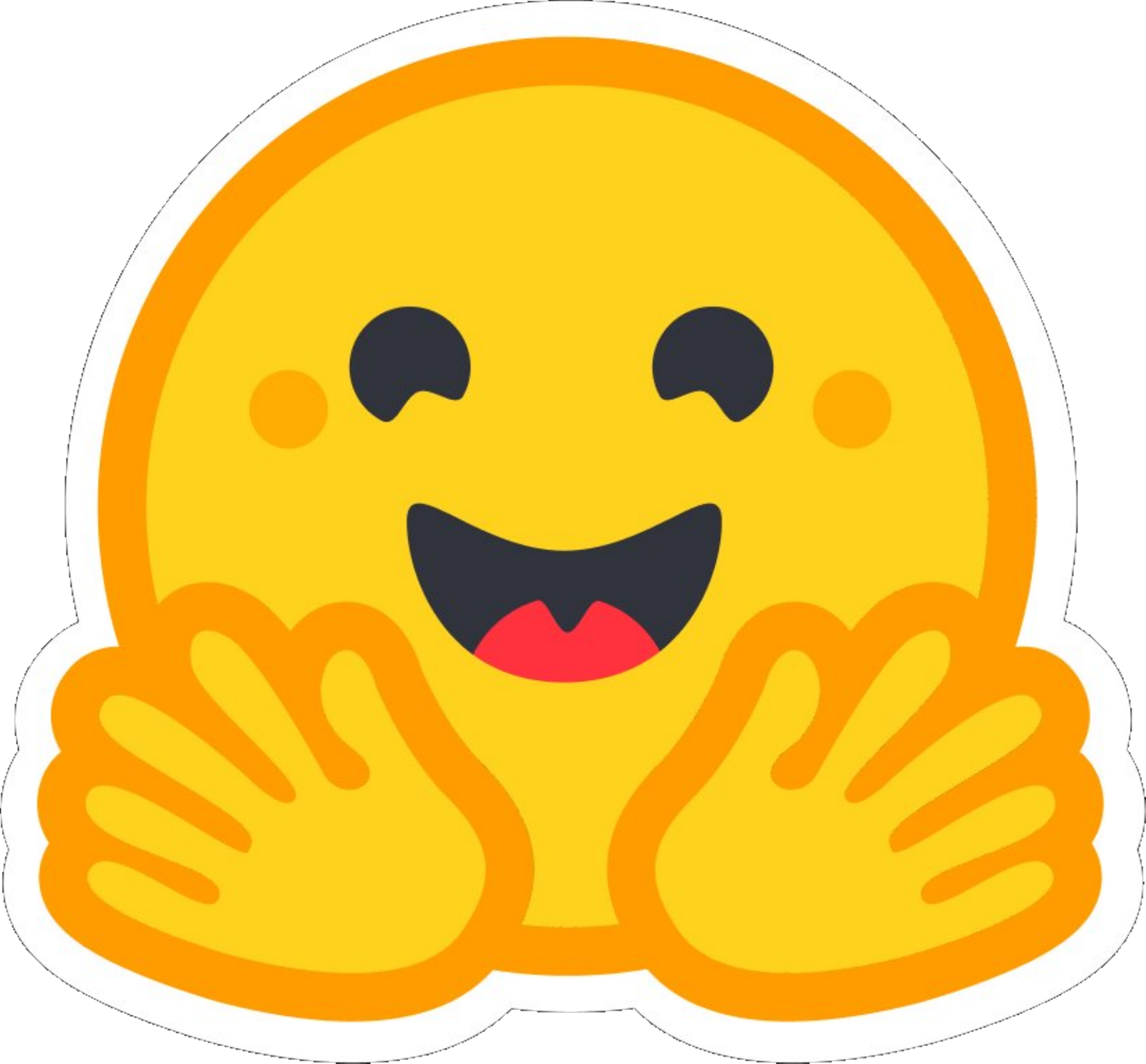}}\xspace}
\renewcommand{\today}{\DTMtoday}

\fancypagestyle{headstyle}{
    \fancyhead[C]{
Data-Efficient Autoregressive-to-Diffusion Language Models via On-Policy Distillation
    }
}

\definecolor{hidden-red}{RGB}{205, 44, 36}
\definecolor{hidden-blue}{RGB}{194,232,247}
\definecolor{hidden-orange}{RGB}{243,202,120}
\definecolor{hidden-green}{RGB}{34,139,34}
\definecolor{hidden-pink}{RGB}{255,245,247}
\definecolor{hidden-black}{RGB}{20,68,106}
\definecolor{purple}{RGB}{144,153,196}
\definecolor{yellow}{RGB}{255,228,123}
\definecolor{hidden-yellow}{RGB}{255,248,203}
\definecolor{tkcolor}{RGB}{224,223,255}
\definecolor{darkblue}{rgb}{0, 0.40, 0.75}

\definecolor{tamuMaroon}{HTML}{500000}
\colorlet{abstractTextColor}{tamuMaroon!15!black}
\newcommand{\abstractstyle}{\color{abstractTextColor}}

\makeatletter
\let\oldabstract\abstract
\let\oldendabstract\endabstract
\renewcommand{\abstract}{\oldabstract\abstractstyle}
\renewcommand{\endabstract}{\oldendabstract}
\makeatother

\hypersetup{colorlinks=true, citecolor=darkblue, linkcolor=darkblue, urlcolor=darkblue}
\tcbset{
  aibox/.style={
    width=\linewidth,
    top=8pt,
    bottom=4pt,
    colback=blue!6!white,
    colframe=black,
    colbacktitle=black,
    enhanced,
    center,
    attach boxed title to top left={yshift=-0.1in,xshift=0.15in},
    boxed title style={boxrule=0pt,colframe=white,},
  }
}

\newtcolorbox{TakeawayBox}[2][]{takeawaybox,title=#2,#1}

\usepackage{mathtools}
\usepackage[utf8]{inputenc} % allow utf-8 input
\usepackage[T1]{fontenc}    % use 8-bit T1 fonts
\usepackage{hyperref}       % hyperlinks
\usepackage{url}            % simple URL typesetting
\usepackage{booktabs}       % professional-quality tables
\usepackage{amsfonts}       % blackboard math symbols
\usepackage{nicefrac}       % compact symbols for 1/2, etc.
\usepackage{microtype}      % microtypography
\usepackage{algorithm}
\usepackage{algorithmic}
\usepackage{amsmath,amssymb}
\usepackage{bbm}
\usepackage{graphicx}
\usepackage[table]{xcolor}
\usepackage{colortbl}
\usepackage{enumitem}
\usepackage{wrapfig}
\usepackage{subcaption}
\usepackage{wrapfig}
\colorlet{tgrad1}{orange}
\colorlet{tgrad2}{orange!90}
\colorlet{tgrad3}{orange!80}
\colorlet{tgrad4}{orange!70}
\colorlet{tgrad6}{green}

\usepackage{xcolor}

\definecolor{mygreen}{HTML}{00A300}

\usepackage[dvipsnames]{xcolor}

\newcommand{\vx}{\mathbf{x}}

\newcommand{\vm}{\mathbf{m}}

\usepackage[capitalise,nameinlink]{cleveref}

\usepackage[normalem]{ulem}

\usepackage[textsize=tiny]{todonotes}      % todos visible

\usepackage{wrapstuff}
\usepackage[dvipsnames]{xcolor}
\usepackage{dsfont}
\usepackage{regexpatch}
\makeatletter
\presetkeys{todonotes}{inline,caption={}}{}
\xpatchcmd{\@todo}{\setkeys{todonotes}{#1}}{\setkeys{todonotes}{inline,#1}}{}{}
\makeatother

\newcommand{\methodname}{\textsc{OPDLM}}
\definecolor{ourscol}{gray}{0.92}

\newcommand{\bb}[1]{\textbf{#1}}  % bold for best in row within scale

\title{Data-Efficient Autoregressive-to-Diffusion Language Models via On-Policy Distillation}

\author{
    Xingyu Su$^{1*}$, 
    Jacob Helwig$^{1*}$, 
    Shubham Parashar$^{1*}$, 
    Atharv Chagi$^{1}$, 
    Lakshmi Jotsna$^{1}$, \\
    \textbf{Degui Zhi}$^{2}$,
    \textbf{James Caverlee}$^{1}$, 
    \textbf{Dileep Kalathil}$^{1,3}$, 
    \textbf{Shuiwang Ji}$^{1}$
    \\
    \normalfont $^1$ Department of Computer Science and Engineering, Texas A\&M University \\
    \normalfont $^2$ Department of Bioinformatics and Systems Medicine, \\
    \normalfont \hspace*{1.2em} University of Texas Health Science Center at Houston \\
    \normalfont $^3$ Department of Electrical and Computer Engineering, Texas A\&M University
}

\begin{document}

\begin{abstract}

  % \vspace{1mm}
  \textbf{\large Abstract:}
  \vspace{1mm}

We study the transformation of autoregressive models (ARLMs) into diffusion language models (DLMs). Rather than pretraining from scratch, prior work replaces the causal attention in ARLMs with bidirectional attention and then trains the resulting model using a DLM objective. However, these approaches incur two distribution shifts. First, transitioning from a next-token prediction objective to a DLM objective can discard knowledge acquired by the ARLM during training. Second, standard DLMs suffer from a train-inference mismatch, as the training loss is defined on randomly masked sequences rather than the  trajectories encountered at inference produced by confidence-based decoding. To address both challenges, we introduce an On-Policy Diffusion Language Model (\methodname) in which 
On-Policy Distillation (OPD) is employed for ARLM-to-DLM transformation. Specifically, \methodname{} is trained via self-OPD, where the student, an ARLM with bidirectional attention, generates its own trajectories, and the teacher, the original frozen ARLM, distills its knowledge by providing target logits on these trajectories. By training directly in an on-policy manner, \methodname{} eliminates the train-inference mismatch in DLMs, while distillation from the original model enhances knowledge retention from the ARLM. Empirical results demonstrate that \methodname{} requires \textbf{15$\times$} to \textbf{7,000$\times$} fewer training tokens with strong performance across a wide variety of tasks. \methodname{} avoids the prohibitive cost of DLM pretraining and positions DLM transformation as a form of ARLM post-training.

  \vspace{5mm}

  % $^{*}$ \textit{Equal Contribution}
  
  % $^{\coloremojicode{2709}}$ \textit{Corresponding Author}

  \vspace{1mm}
  \vspace{6mm}

  % \coloremojicode{1F4C5} \textbf{Date}: 

  % \coloremojicode{1F3E0} \textbf{Homepage}: https://agenticscience.github.io/

  % \github{} \textbf{Github Repository}: https://github.com/AgenticScience/AgenticScience.github.io

  % \coloremojicode{1F4E7} \textbf{Correspondence}: Siqi Sun, \href{}{siqisun@fudan.edu.cn}

  % \coloremojicode{1F4E7} \textbf{Equal contribution}: Jiaqi Wei, Yuejin Yang, Xiang Zhang, Yuhan Chen

  % \textbf{Date}: August 6, 2025

 % \coloremojicode{1F4C5} \textbf{Date}: August, 2025

   \textbf {*: These authors contributed equally}
   
   % \textbf {$\dag$: These authors jointly led the project}

  \vspace{3mm}

  \coloremojicode{1F3E0} \textbf{Homepage}: 
  \href{https://opdlm.vercel.app/}{https://opdlm.vercel.app}

  \github{} \textbf{Github Repository}: 
  \href{https://github.com/divelab/OPDLM}{https://github.com/divelab/OPDLM}

  \huggingface{} \textbf{HuggingFace Space}: 
  \href{https://huggingface.co/collections/divelab/opdlm}
  {https://huggingface.co/collections/divelab/opdlm}
  
  % \coloremojicode{1F4E7} \textbf{Correspondence}: \href{}{siqisun@fudan.edu.cn}

    % \coloremojicode{1F4E7} \textbf{Contact}: \href{}{ \textbraceleft weijiaqi, yangyuejin\textbraceright@pjlab.org.cn}

    \vspace*{0.2in}

\end{abstract}

\maketitle

% \vspace{3mm}
\pagestyle{headstyle}
\thispagestyle{empty}

\newpage
\vspace{2em}
{\renewcommand{\baselinestretch}{0.975}\selectfont
\tableofcontents
}
\newpage
\section{Introduction}

\begin{figure}[t]
  \centering
  \includegraphics[width=1\textwidth]{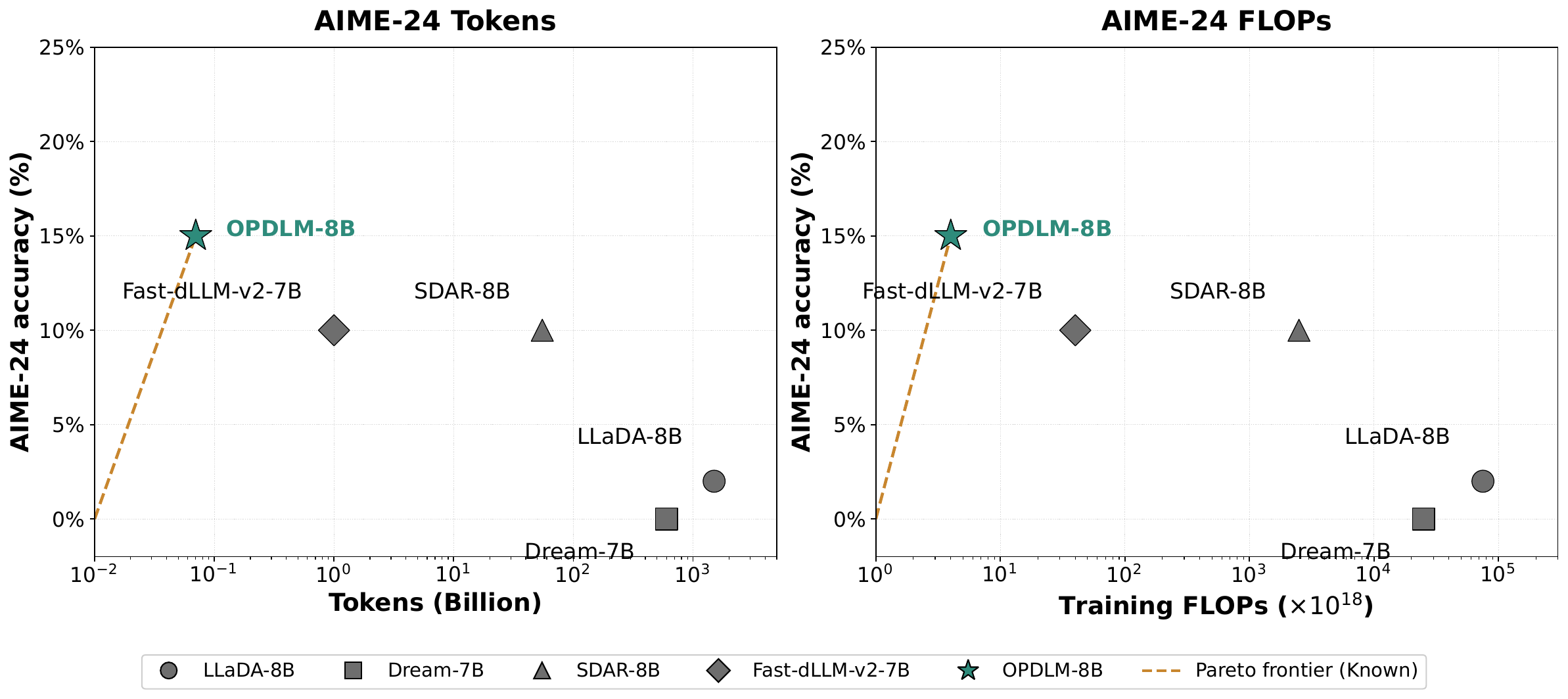}
  \caption{OPDLM establishes a new pareto frontier for AIME-24~\citep{aime24}.  Notably, training OPDLM-8B requires only $0.066\text{B}$ training tokens and $4.2 \times 10^{18}$ FLOPs, which is 15$\times$ to 7,000$\times$ less as compared to established DLMs obtained from ARLMs~\citep{ye2025dream7bdiffusionlarge, cheng2025sdar, wu2025fast}, demonstrating compute and data efficiency.}
  \label{fig:pareto}
\end{figure}

Diffusion Language Models (DLMs) extend diffusion-based generative modeling
from domains such as images, video, robotics, and biology
~\citep{ho2020denoising, song2019generative, dhariwal2021diffusion,
ho2022video, pearce2023imitating, hayes2024simulating}
to text~\citep{austin2021structured}. A prominent class of DLMs, Masked
Diffusion Language Models (MDLMs)~\citep{sahoo2024simple}, has recently shown
competitive performance with autoregressive language models (ARLMs) on
arithmetic, coding, and general reasoning tasks~\citep{ye2025dream7bdiffusionlarge,
nie2025large}, while offering the potential advantage of decoding multiple
tokens per step. Unlike ARLMs, which are trained by next-token prediction,
MDLMs are pretrained by masking tokens in training sequences and minimizing a
negative evidence lower bound (NELBO)~\citep{sahoo2024simple}. Since
pretraining MDLMs from scratch can require trillions of tokens
~\citep{nie2025large, arriola2025block}, recent work instead converts
pretrained ARLM checkpoints into DLMs by replacing causal attention with
bidirectional attention and continuing training with a diffusion objective,
reducing the training requirement to billions of tokens
~\citep{ye2025dream7bdiffusionlarge, cheng2025sdar, wu2025fast}. However,
this conversion process still leaves two mismatches that limit training
efficiency.

The first is a knowledge-retention mismatch. A pretrained ARLM contains
substantial knowledge acquired through next-token prediction, but replacing
its attention structure and optimizing an MDLM objective can weaken or discard
some of that knowledge. The second is a training--inference state mismatch:
MDLMs are trained on uniformly random masked states, whereas inference follows
model- and sampler-induced reverse unmasking trajectories, typically using
confidence-guided decoding heuristics~\citep{nie2025large}. These trajectories
are not directly represented by the standard forward-masking training
distribution. This motivates two questions: \emph{How can we retain knowledge
from the original ARLM during conversion into a DLM?} and \emph{how can we
train the converted model directly on the states it visits during diffusion
inference?}

In autoregressive settings, On-Policy Distillation (OPD)~\citep{agarwal2024onpolicy}
offers a useful template for these goals. OPD supervises a student on states
sampled from its own generation process, reducing exposure bias while
distilling token-level knowledge from a teacher model. However, applying OPD to
ARLM-to-DLM conversion is not straightforward, as standard OPD assumes the teacher and student share the same state space. For a DLM student,
these states are partially masked diffusion states; therefore, a direct application would require a trained DLM teacher to provide targets on such
states. This is precisely the requirement we aim to avoid.

In this work, we introduce On-Policy Diffusion Language Models
(\methodname{}), an on-policy method for converting ARLMs into DLMs. The DLM
student is initialized from the ARLM checkpoint, samples its own reverse
diffusion trajectories, and is trained on those trajectories using token-level
targets from the original frozen ARLM. This combines on-policy training of the
DLM student with knowledge distillation from the ARLM teacher, avoiding the need
for a separately pretrained DLM teacher.

Our contributions are as follows. First, we formulate ARLM-to-DLM conversion as
a post-training problem and show that DLMs can be obtained without the
computationally intensive DLM pretraining stage. Second, we demonstrate strong
data efficiency: \methodname{} uses 15$\times$ to 7{,}000$\times$ fewer
training tokens than established DLM baselines while remaining competitive
across model scales and task categories (see~\cref{fig:pareto}). Third, we
provide evidence that ARLM supervision supports knowledge retention during
conversion, including zero-shot behavior on capabilities not explicitly targeted
during conversion, such as multilingual generation and extended thinking.
Finally, we show that the same framework can produce both general-purpose and
task-specialized DLMs. We release our data, code, and model checkpoints to
support future work.

\begin{figure}
    \centering
    \includegraphics[width=0.99\linewidth]{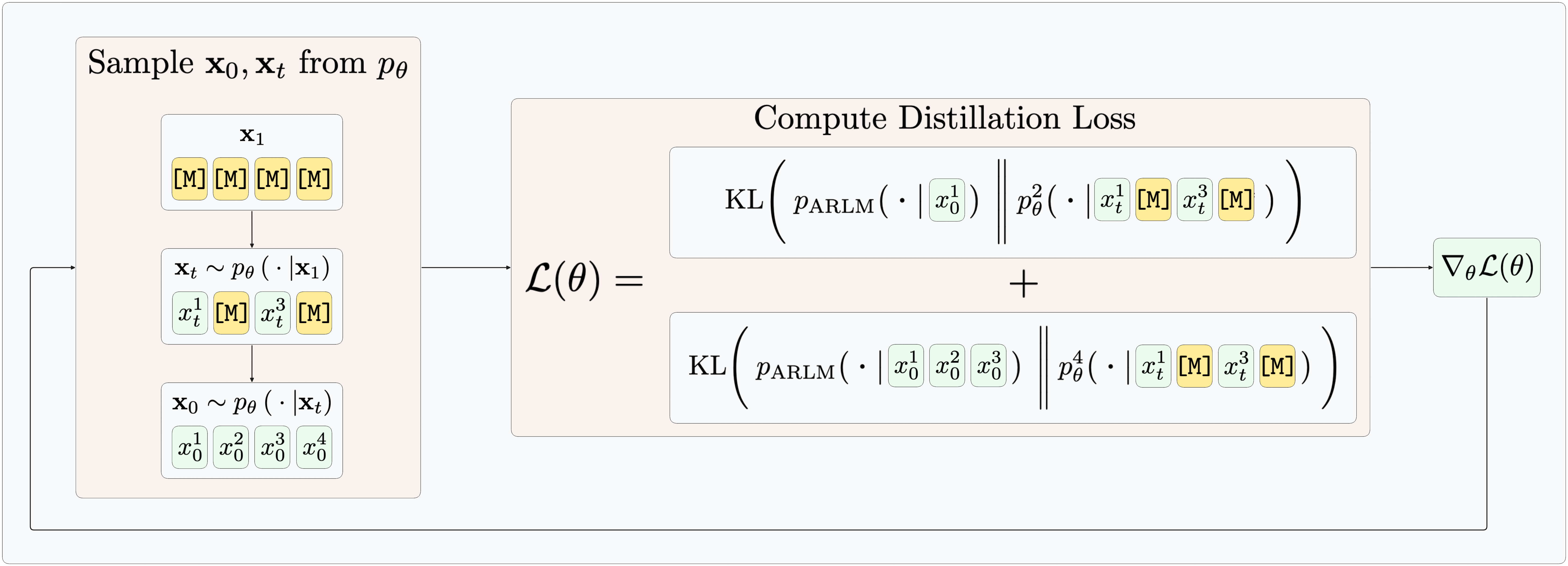}
    \caption{OPDLM training framework. At each training step, we sample a reverse decoding trajectory from the student DLM $p_\theta$. From this trajectory, we select one partially denoised state $\vx_t$ and compute a distillation loss over its masked tokens. For each masked position $i$, the loss aligns the student distribution $p_\theta^i$ with the frozen ARLM teacher distribution conditioned on the corresponding causal prefix of the terminal sequence $\vx_0$. This trains the DLM on its own inference-time states while transferring knowledge from the original ARLM. For simplicity, the figure shows a single block of size 4, while in practice, trajectories contain multiple blocks.}
    \label{fig:opdlm}
\end{figure}

\section{Background and Related Works}\label{sec:related}

\textbf{Masked Diffusion Language Models} (MDLMs) have emerged as an alternative to autoregressive models (ARLMs) for text generation~\citep{sahoo2024simple, nie2025large}. Structurally, MDLMs fall into two distinct paradigms: global diffusion across all masked positions~\citep{nie2025large, sahoo2024simple} and localized diffusion within a block-wise causal framework~\citep{arriola2025block}. DLMs enable more flexible decoding and demonstrate superior training efficiency with improved convergence on perplexity~\citep{Nietal2025, Gaoetal2025}. Despite these advantages, pretraining these models from scratch requires trillion-scale tokens~\citep{nie2025large}.

\textbf{ARLM to DLM Conversion} is an efficient paradigm for pretraining DLMs~\citep{ye2025dream7bdiffusionlarge, wu2025fast, cheng2025sdar, nbdiff, efficientdlm, zhou2026dllm, LLaDA2}. Such pretraining reduces the required training data from trillions of tokens to billions, but the current conversion process is largely treated as a phase of continued pretraining~\citep{Xieetal2024, mehta2026breaking} in which the next-token prediction loss of ARLMs is replaced by the standard DLM loss with bidirectional attention. The adaptation of an ARLM into a DLM has not been explored using knowledge distillation~\citep{Hintonetal2015}. %, primarily because the teacher (ARLM) and student (DLM) factorize text distributions differently.

% Resolving this severe objective mismatch necessitates an asymmetric distillation approach~\citep{wu2022contextual} to distill knowledge across distinct generative paradigms.

\textbf{On-Policy Distillation} (OPD) methods train a student model on trajectories sampled directly from its own policy, while a teacher model provides token-level supervision via KL divergence~\citep{agarwal2024onpolicy, qwen3, deepseekv4}. Unlike standard knowledge distillation~\citep{Hintonetal2015}, querying the teacher on these student-generated states has been shown to mitigate the distribution shift arising from learning with teacher generations~\citep{dagger}. Recently, self-OPD for ARLMs~\citep{zhao2026self, hubotter2026sdpo, shenfeld2026sdft} has demonstrated that a single model can serve as its own teacher by leveraging privileged information to supervise a student conditioned on a strictly weaker context.
In this work, we investigate whether OPD can be extended to train a student DLM with supervision from an ARLM teacher.

\section{Preliminaries}

% In this section, we provide an overview of diffusion language models (DLMs) and  On-Policy Distillation (OPD) for autoregressive language models (ARLMs).
In this section, we review masked diffusion language models (MDLMs) and
On-Policy Distillation (OPD) for autoregressive language models (ARLMs).

\textbf{Masked Diffusion Language Models (MDLMs).} MDLMs define a forward masking process that corrupts a clean sequence\footnote{Unless otherwise stated, we write $\vx$ for the sequence being denoised.
For conditional generation, any prompt is treated as fixed observed context
and omitted from the notation.}
$\vx_0\sim p_{\mathrm{data}}$, with $|\vx_0|\le L$. For diffusion time
$t\in[0,1]$, the forward process independently replaces each token with the
\textsc{mask} token $\vm$:
\[
    q^{t\mid0}(x_t^i \mid x_0^i)
    =
    \alpha_t \delta_{x_0^i}(x_t^i)
    +
    (1-\alpha_t)\delta_{\vm}(x_t^i),
    \qquad
    \alpha_t = 1-t .
\]
We use \(q^{t\mid0}(\cdot\mid\vx_0)\) to denote the corresponding product distribution over sequence-level corruptions. 
Thus, sampling $t\sim\mathcal U(0,1)$ corresponds to masking each token in
$\vx_0$ independently with probability $t$~\citep{nie2025large,sahoo2024simple}. 
For later comparison with on-policy training, it is useful to view this
forward process as a distribution over trajectories. 
Given $\vx_0$, let
$\tau^{\mathrm{fwd}} \sim \Gamma^{\mathrm{fwd}}(\cdot\mid\vx_0)$
denote a forward masking trajectory, viewed as a map from diffusion time to
corrupted sequences, whose time-$t$ marginal satisfies
$\tau^{\mathrm{fwd}}(t)\sim q^{t\mid0}(\cdot\mid\vx_0)$.
% Given $\vx_0$, let
% $\tau^{\mathrm{fwd}}=\{X_t\}_{t\in[0,1]}
% \sim \Gamma^{\mathrm{fwd}}(\cdot\mid\vx_0)$ denote a forward masking
% trajectory whose time-$t$ marginal satisfies
% $X_t(\tau^{\mathrm{fwd}})\sim q^{t\mid0}(\cdot\mid\vx_0)$.
% Standard MDLM training can then be viewed as sampling a trajectory
% $\tau^{\mathrm{fwd}}$, selecting a time $t\sim\mathcal U(0,1)$, and setting
% $\vx_t = X_t(\tau^{\mathrm{fwd}})$. 
Standard MDLM training can then be viewed as sampling a trajectory
\(\tau^{\mathrm{fwd}}\), selecting a time \(t\sim\mathcal U(0,1)\), and setting
\(\vx_t=\tau^{\mathrm{fwd}}(t)\).
Given this corrupted sequence, the model
is trained to recover the original tokens at masked positions as
\begin{equation}
\label{eq:mdlm-nelbo}
    \mathcal{L}_{\mathrm{MDLM}}(\theta)
    =
    - \mathbb{E}_{\vx_0 \sim p_{\mathrm{data}},\,
    t \sim \mathcal{U}(0,1),\,
    \vx_t \sim q^{t \mid 0}(\cdot\mid\vx_0)}
    \left[
    \frac{1}{t}
    \sum_{i\in\mathcal{M}(\vx_t)}
    \log p_\theta(x_0^i\mid \vx_t)
    \right],
\end{equation}
where $\mathcal{M}(\vx)=\{i:x^{i}=\vm\}$ is the set of masked positions and the factor $1/t$ is the diffusion time weight induced by the linear
schedule. Following~\citep{arriola2025block}, we use \textbf{block diffusion}, which
partitions the sequence into $B$ non-overlapping contiguous blocks. For block
$b$, the model conditions on the clean prefix $\vx_0^{<b}$ and the corrupted
active block $\vx_t^b$ as
\begin{equation}
\label{eq:mdlm-block-nelbo}
     \mathcal{L}_{\mathrm{block}}(\theta)
     =
     - \mathbb{E}_{\vx_0 \sim p_{\mathrm{data}},\,
     t \sim \mathcal{U}(0,1),\,
     \vx_t \sim q^{t \mid 0}(\cdot\mid\vx_0)}
     \left[
     \sum_{b=1}^{B}
\frac1t\sum_{i\in\mathcal{M}(\vx^b_t)}
     \log p_\theta(x_0^{b,i}\mid \vx_0^{<b},\vx_t^b)
     \right].
\end{equation}

During inference, a new sequence is generated through a reverse unmasking
procedure induced jointly by the model and sampler. Starting from an initially
masked sequence, the sampler iteratively queries the model on the current
partially denoised state, \(p_\theta(\cdot\mid \vx_{t_k})\), and selects and
updates a subset of masked positions according to a decoding rule, typically
based on model confidence. This process produces a reverse decoding trajectory
$\tau \coloneq (\vx_{t_T}, \vx_{t_{T-1}}, \ldots, \vx_{t_0})$,
where $\vx_{t_T}$ is the initial masked sequence, $\vx_{t_0}$ is the terminal
generated sequence, and the number of steps $T$ and realized effective
denoising times $t_T,t_{T-1},\ldots,t_0$ are induced by the realized
interaction between the model and sampler. We view
$\tau$ as a map on these realized times, so that $\tau(t_k)=\vx_{t_k}$. For a fixed sampler, we denote the resulting reverse trajectory distribution by
$\tau \sim \Gamma_{\theta}^{\mathrm{rev}}$.

The distinction between forward random-masking states during training and
reverse sampler-induced states during inference highlights the
training--inference mismatch in standard MDLM training. During training, the loss is evaluated on states obtained by sampling
$\vx_0\sim p_{\mathrm{data}}$, $t\sim\mathcal U(0,1)$, and
$\vx_t\sim q^{t\mid0}(\cdot\mid\vx_0)$, whereas during inference the model
encounters intermediate states $\tau(t_k)$ along trajectories sampled from
$\Gamma_{\theta}^{\mathrm{rev}}$. 
Thus, the training state distribution need
not match the inference state distribution, which can make learning less
efficient. 

\textbf{On-Policy Distillation (OPD).}
OPD~\citep{agarwal2024onpolicy} reduces the analogous exposure bias in
autoregressive models by supervising a student on prefixes sampled from its own
generation process. The student first samples an on-policy
sequence $\hat{\vx}\sim p_\theta$. The teacher $p_\phi$ is then queried on the
causal prefixes of this same sequence, producing token-level targets for the
student as
\begin{equation}
\label{eq:opd-arlm}
    \mathcal{L}_{\mathrm{OPD}}(\theta)
    =
    \mathbb{E}_{\hat{\vx}\sim p_\theta}
    \left[
    \frac{1}{|\hat{\vx}|}
    \sum_{i=1}^{|\hat{\vx}|}
    D_{\mathrm{KL}}
    \left(
    p_\phi(\cdot\mid \hat{\vx}_{<i})
    \;\middle\|\;
    p_\theta(\cdot\mid \hat{\vx}_{<i})
    \right)
    \right].
\end{equation}
The distinction from supervised distillation is the state distribution on which the student and teacher distributions are compared: OPD trains on
prefixes induced by the student's own rollout rather than on prefixes from a
fixed dataset or from teacher-generated responses.

% This perspective suggests an analogous goal for DLM conversion: supervise the
% student on the partially masked states it visits during reverse decoding,
% rather than on states produced by forward random masking. A direct analogue,
% however, would require a trained DLM teacher that can provide targets on such
% partially masked states. OPDLM avoids this requirement by using the original
% frozen ARLM as the teacher while sampling training states from the student
% DLM's reverse trajectories.

\section{On-Policy Diffusion Language Models}

We propose On-Policy Diffusion Language Models (OPDLM), an on-policy method
for efficiently converting an ARLM into a DLM. We begin by writing block
diffusion training as a general objective over trajectory distributions, which
will allow us to compare standard forward-masking training with OPDLM's
reverse-trajectory training. Let $\Gamma$ denote a distribution over trajectories $\tau\sim\Gamma$, viewed
as maps from times to sequence states, and
let $\nu(\cdot\mid\tau)$ denote a state-selection distribution over times in
the trajectory. For a sampled trajectory and selected time
$t\sim\nu(\cdot\mid\tau)$, we write $\vx_t=\tau(t)$ for the selected
intermediate state and $\vx_0=\tau(0)$ for the terminal sequence. We define the generalized
block diffusion objective as
\begin{equation}
\label{eqn:general_abstract}
\mathcal{L}_{\Gamma,\nu,\phi}(\theta)
=
\mathbb{E}_{\substack{
\tau\sim\Gamma,\,
t\sim\nu(\cdot\mid\tau)
}}
\left[
\sum_{b=1}^{B}
w(t)
\sum_{i\in\mathcal{M}(\vx_t^b)}
D_{\mathrm{KL}}
\left(
\phi^{b,i}(\cdot\mid \vx_0)
\;\middle\|\;
p_\theta^{b,i}(\cdot\mid \vx_0^{<b},\vx_t^b)
\right)
\right].
\end{equation}
Here, $\Gamma$ determines the source of training trajectories,
$\nu$ determines which state along each trajectory is supervised, $w(t)$ is a
time-dependent weight, $\phi^{b,i}$ is the target distribution at position $i$
in block $b$, and $p_\theta^{b,i}$ is the DLM student's predictive distribution
at the same position.

This formulation recovers the standard block-diffusion loss in
\cref{eq:mdlm-block-nelbo} as a special case. 
Choose $\Gamma$ to be the trajectory distribution induced by first sampling
$\vx_0\sim p_{\mathrm{data}}$ and then sampling
$\tau\sim\Gamma^{\mathrm{fwd}}(\cdot\mid\vx_0)$, where
$\tau(t)\sim q^{t\mid0}(\cdot\mid\vx_0)$ for every $t$. Taking
$\nu=\mathcal U(0,1)$, $w(t)=1/t$ for the linear schedule
$\alpha_t=1-t$, and hard-label targets
$\phi^{b,i}(\cdot\mid\vx_0)=\delta_{x_0^{b,i}}$ recovers
\cref{eq:mdlm-block-nelbo}, since
$D_{\mathrm{KL}}\left(
\delta_{x_0^{b,i}}
\;\middle\|\;
p_\theta^{b,i}(\cdot\mid \vx_0^{<b},\vx_t^b)
\right)
=
-\log p_\theta^{b,i}(x_0^{b,i}\mid \vx_0^{<b},\vx_t^b)$.
This trajectory-level view isolates the source of the training--inference
mismatch in standard block diffusion. Training uses states induced by the data-marginal forward random-masking
trajectory distribution, obtained by sampling $\vx_0\sim p_{\mathrm{data}}$
and then $\tau\sim\Gamma^{\mathrm{fwd}}(\cdot\mid\vx_0)$, whereas inference
uses states induced by the reverse decoding trajectory distribution
$\Gamma_{\theta}^{\mathrm{rev}}$.
In this framework, directly reducing the mismatch amounts to changing the
training trajectory distribution.

\subsection{Addressing Training--Inference Mismatch}

Guided by the trajectory objective in~\cref{eqn:general_abstract}, OPDLM
replaces the forward masking trajectory distribution with the student's
sampler-induced reverse trajectory distribution. Specifically, we sample a
reverse trajectory
$
    \tau = (\vx_{t_T},\vx_{t_{T-1}},\ldots,\vx_{t_0})
    \sim \Gamma_{\theta}^{\mathrm{rev}},
$
where $\Gamma_{\theta}^{\mathrm{rev}}$ is induced by the student DLM and fixed
sampler, and $t_T,\ldots,t_0$ are the realized effective denoising
times along the sampled trajectory, with $t_T=1$ and $t_0=0$. Here,
$\vx_0=\tau(t_0)$ denotes the terminal generated sequence of the sampled
reverse trajectory. We then sample an index
$k\sim\mathcal U(\{1,\ldots,T\})$, set $t=t_k$, and take
$\vx_t=\tau(t_k)$.
Equivalently, this samples
$t$ from the state-selection distribution
$\nu_{\mathrm{rev}}(\cdot\mid\tau)$ that is uniform over the realized
non-terminal times $\{t_1,\ldots,t_T\}$. 
Substituting this choice of $\Gamma$ and $\nu_{\mathrm{rev}}$ into
\cref{eqn:general_abstract} gives the on-policy block diffusion objective
\begin{equation}
\label{eqn:train_inf}
\mathcal{L}_{\mathrm{on}}(\theta)
=
\mathbb{E}_{\substack{
\tau\sim\Gamma_{\theta}^{\mathrm{rev}},\,
t\sim\nu_{\mathrm{rev}}(\cdot\mid\tau)
}}
\left[
\sum_{b=1}^{B}
w(t)
\sum_{i\in\mathcal M(\vx_t^b)}
D_{\mathrm{KL}}
\left(
\phi^{b,i}(\cdot\mid \vx_0)
\;\middle\|\;
p_\theta^{b,i}(\cdot\mid \vx_0^{<b},\vx_t^b)
\right)
\right].
\end{equation}
This objective is on-policy because the supervised states are sampled from the
same reverse decoding process used at inference. It still leaves open the
choice of target distribution $\phi^{b,i}$ at each masked position. A trained DLM
teacher could provide such targets, but would defeat our goal of converting an
ARLM into a DLM without first training a separate DLM. We therefore define
$\phi^{b,i}$ using the original frozen ARLM, and $p_{\theta}$ as the DLM student initialized from the same ARLM checkpoint but with blockwise-causal attention.

\subsection{ARLM Supervision for On-Policy States}

\begin{algorithm}[t]
\small
\caption{OPDLM Training Step}
\label{alg:opdlm}
\begin{algorithmic}[1]
   \STATE {\bfseries Input:} frozen ARLM teacher $p_{\mathrm{ARLM}}$; student DLM $p_\theta$; sampler $\mathcal S$
   \STATE {\bfseries // On-policy reverse rollout}
   \STATE Sample reverse trajectory
   $\tau=(\vx_{t_T},\vx_{t_{T-1}},\ldots,\vx_{t_0})
   \sim \Gamma_{\theta}^{\mathrm{rev}}$
   \STATE Set terminal sequence $\vx_0 \gets \tau(t_0)$
   \STATE Sample non-terminal trajectory index $k\sim\mathcal U(\{1,\ldots,T\})$
   \STATE Set training time $t\gets t_k$ and on-policy state $\vx_t\gets \tau(t_k)$
   \STATE Initialize loss $\mathcal L\gets 0$

   \STATE {\bfseries // ARLM-supervised distillation}
   \FOR{each block $b=1,\ldots,B$}
      \FOR{each masked position $i\in\mathcal M(\vx_t^b)$}
         \STATE Set teacher distribution
         $p_{\mathrm T}\gets p_{\mathrm{ARLM}}(\cdot\mid \vx_0^{<b},\vx_0^{b,<i})$
         \STATE Set student distribution $p_{\mathrm S}\gets p_\theta^{b,i}(\cdot\mid \vx_0^{<b},\vx_t^b)$
         \STATE Update loss
         $\mathcal L\gets \mathcal L + D_{\mathrm{KL}}(p_{\mathrm T}\parallel p_{\mathrm S})$
      \ENDFOR
   \ENDFOR
   \STATE {\bfseries Output:} $\mathcal L$
\end{algorithmic}
\end{algorithm}
To define the target
distribution $\phi^{b,i}$ for a partially masked diffusion state, a trained DLM
teacher could be queried directly on $\vx_t$, but OPDLM instead uses the frozen
ARLM teacher. This choice serves two purposes: it avoids requiring a separately
trained DLM teacher, and it transfers knowledge from the original ARLM through
token-level distribution matching.

The main challenge is that the student DLM predicts from a partially masked
block $\vx_t^b$, whereas the ARLM teacher is defined over unmasked causal prefixes.
We resolve this structural mismatch by using the terminal sequence
$\vx_0=\tau(t_0)$ from the student's reverse trajectory to construct the
teacher prefix. 
For each masked position $i$ in block $b$, we define the target distribution by querying the teacher on the unmasked causal prefix:
$\phi^{b,i}(\cdot\mid \vx_0)
=
p_{\mathrm{ARLM}}(\cdot\mid \vx_0^{<b},\vx_0^{b,<i})$, 
where $p_{\mathrm{ARLM}}$ denotes the original frozen ARLM.
% \]
Substituting this target distribution into~\cref{eqn:train_inf} gives the
OPDLM objective as
\begin{equation*}
\label{eqn:opdlm}
\resizebox{\linewidth}{!}{$\displaystyle
\mathcal{L}_{\mathrm{OPDLM}}(\theta)
=
\mathbb{E}_{\substack{
\tau\sim\Gamma_{\theta}^{\mathrm{rev}},\,
t\sim\nu_{\mathrm{rev}}(\cdot\mid\tau)
}}
\left[
\sum_{b=1}^{B}
w(t)
\sum_{i\in\mathcal M(\vx_t^b)}
D_{\mathrm{KL}}
\left(
p_{\mathrm{ARLM}}(\cdot\mid \vx_0^{<b},\vx_0^{b,<i})
\;\middle\|\;
p_\theta^{b,i}(\cdot\mid \vx_0^{<b},\vx_t^b)
\right)
\right].
$}
\end{equation*}

The student is therefore trained on its own reverse-trajectory states, with supervision provided by the predictive distribution of the original ARLM. Because the target is the ARLM's full token-level predictive distribution at each supervised position, this objective encourages retention of knowledge from the original model. For the loss weighting, we follow~\citet{zhou2026dllm} and set $w(t)=1$; although related work has explored alternative schemes~\citep{ye2025dream7bdiffusionlarge, efficientdlm}, we find this simple uniform weighting sufficient. The training procedure is visualized in~\cref{fig:opdlm} and detailed step-by-step in~\cref{alg:opdlm}.

\subsection{Rollout-Length Curriculum}

Although OPDLM trains on the student's own reverse trajectories, these
trajectories can be unstable immediately after ARLM-to-DLM conversion. The
student inherits the ARLM weights, but is now queried in a different mode:
masked-token embeddings appear in the input, attention is bidirectional within
each block, and predictions are made at masked positions rather than by
standard next-token prediction. As a result, long early rollouts may produce
low-quality terminal sequences, causing the ARLM teacher to provide targets on
incoherent causal prefixes. We stabilize training using a curriculum learning inspired strategy~\cite{bengio2009curriculum, parashar2026curriculum}, progressively increasing the length of the rollout as the student improves. Let $L_{\min}$ and $L_{\max}$ denote the minimum and maximum sequence lengths used for on-policy generation.
At training step $s$, we set
\[
    L_s
    =
    \min\left(
    L_{\max},
    L_{\min}
    +
    \left\lfloor
    \frac{s}{S_{\mathrm{warm}}}
    (L_{\max}-L_{\min})
    \right\rfloor
    \right),
\]
where $S_{\mathrm{warm}}$ is the number of warmup steps. Early in training,
this restricts optimization to shorter rollouts, where the student 
learns to match the ARLM teacher on contexts primarily comprised of the prompt. As training progresses,
the curriculum gradually exposes the student to longer trajectories while
avoiding the cost of long, low-quality early rollouts that provide weak training signal.

\colorlet{grad1}{green!10}
\colorlet{grad2}{green!20}
\colorlet{grad3}{green!30}
\colorlet{grad4}{green!50}
\colorlet{grad5}{green!50}
\colorlet{grad6}{green!90}
\colorlet{grad7}{gray!15}

\definecolor{mygreen}{HTML}{00D100}
\begin{table*}[t]
\centering
\footnotesize
\setlength{\tabcolsep}{4pt}
% \caption{Main results across 4B and 8B scales. \textbf{OPDLM} is distilled with a non-thinking objective. OPDLM gets strong performances while using \textbf{$\sim$700$\times$ fewer tokens} (0.075B vs.\ 55B). Best result within each scale is in \textbf{bold}. \shubham{Pareto optimal}}
\caption{\textbf{OPDLM 4B} and \textbf{8B} achieves competitive performance across general knowledge, mathematics, and code generation benchmarks while requiring dramatically fewer training resources - as little as \textbf{0.075B tokens}, representing up to a \textbf{15$\times$ to 7,000$\times$ reduction in training tokens} when compared to the other baselines. Cell shading in the Tokens and FLOPs rows reflects training efficiency, darker green indicates fewer tokens/FLOPs consumed.}
\label{tab:main}
\resizebox{\linewidth}{!}{%
\begin{tabular}{l rr rrrrr}
\toprule
 & \multicolumn{2}{c}{\textbf{4B Scale}} & \multicolumn{5}{c}{\textbf{8B Scale}} \\
\cmidrule(lr){2-3} \cmidrule(lr){4-8}
\textbf{Benchmark} 
  & SDAR-4B & \cellcolor{grad1}\textbf{OPDLM-4B}
  & LLaDA-8B & Dream-7B  & SDAR-8B & Fast-dLLM-v2-7B & \cellcolor{grad1}\textbf{OPDLM-8B} \\
\midrule
\textit{Tokens} $\downarrow$
  & \cellcolor{grad3}{55B} & \cellcolor{mygreen}\bb{0.076B}
  & \cellcolor{white}1500B & \cellcolor{grad2}580B & \cellcolor{grad3}55B & \cellcolor{grad4}1B & \cellcolor{mygreen}\bb{0.066B} \\
\textit{FLOPs (1e18)} $\downarrow$
  & \cellcolor{grad3}{1320} & \cellcolor{mygreen}\bb{2.4}
  & \cellcolor{white}72000 & \cellcolor{grad2}24360 & \cellcolor{grad3}2640 & \cellcolor{grad4}42 & \cellcolor{mygreen}\bb{4.2} \\
\midrule
\multicolumn{8}{c}{\textit{General Knowledge \& Instruction Following}} \\
\midrule
MMLU         & 74.9 & \cellcolor{grad1}65.5 & 65.5 & 67.0 &  78.6 & 66.6 & \cellcolor{grad1}70.9 \\
MMLU-Pro     & 50.9 & \cellcolor{grad1}46.3     & 37.0 &  43.3    &  56.9 & 41.5 & \cellcolor{grad1}53.7 \\
% MMLU-Redux   & 75.3 & \cellcolor{ourscol}69.7 &      &      &  78.9 & 71.1 & \cellcolor{ourscol}74.0 \\
GPQA-Diamond & 33.0 & \cellcolor{grad1}29.1     & 31.8 &  32.1    &  40.2 & 27.3 & \cellcolor{grad1}36.1 \\
IFEval       & 56.6 & \cellcolor{grad1}53.8 & 59.9 & 62.5 & 61.4 & 65.4 & \cellcolor{grad1}50.1 \\
CEval        & 62.9 & \cellcolor{grad1}66.9     &   -  &   -  &  70.2 & 70.3 & \cellcolor{grad1}73.3 \\
LiveBench    & 25.3 & \cellcolor{grad1}27.8     &   -  &   -  &   28.6 & 9.5 & \cellcolor{grad1}25.8 \\
\midrule
\multicolumn{8}{c}{\textit{Mathematics \& Reasoning}} \\
\midrule
GSM8K        & 89.9 & \cellcolor{grad1}87.6 & 78.6 & 81.0 &  91.3 & 83.7 & \cellcolor{grad1}87.1 \\
MATH-500     & 72.8 & \cellcolor{grad1}72.8 & 26.6 &  39.2    &  78.6 & 65.6 & \cellcolor{grad1}71.2 \\
AIME-24      & 10.0 & \cellcolor{grad1}14.4 & 2.1 & 0.0 & 10.0 & 10.0 & \cellcolor{grad1}14.7 \\
AIME-25      &  7.5 & \cellcolor{grad1}12.6 & 0.4 & 0.0 &  10.0 & 0.0 &  \cellcolor{grad1}12.4 \\
LMB-Hard     &  6.9 & \cellcolor{grad1}11.1 &   -  & - &  8.9 & 8.9 & \cellcolor{grad1}20.0 \\
ZebraLogic   &  6.3 & \cellcolor{grad1}10.5 &   -  & - &  7.8 &  3.5 & \cellcolor{grad1}12.9 \\
\midrule
\multicolumn{8}{c}{\textit{Code Generation}} \\
\midrule
HumanEval-base & 76.8 & \cellcolor{grad1}56.1 & 35.4 & 57.9 &  82.3 & 63.4 & \cellcolor{grad1}59.8 \\
MBPP-base      & 80.7 & \cellcolor{grad1}57.7 & 31.5 & 68.3 &  79.6 & 63.0 & \cellcolor{grad1}48.7 \\
LCB-v6         & 12.6 & \cellcolor{grad1}10.4     &   -  &  -   &   14.5 & 9.7 & \cellcolor{grad1}9.7 \\
Codeforces     &  4.0 & \cellcolor{grad1}5.0     &   -  &  -   &    5.8 & 5.0 & \cellcolor{grad1}3.5 \\
\bottomrule
\end{tabular}}
\end{table*}

\section{Experiments}
\label{sec:experiments}

In this section, we present experimental results validating the benefits of On-Policy Diffusion Language Models (OPDLMs). We first outline our experimental setup (\cref{sec:setup}) and then present our main results on general-purpose benchmarks (\cref{sec:main_results}). We further conduct an ablation study to disentangle the contributions of OPDLM's design choices (\cref{sec:sft}), evaluate inference-time efficiency through multi-token decoding (\cref{sec:inference}), and demonstrate OPDLM's effectiveness as a specialized DLM (\cref{sec:posttrain}).

\subsection{Setup}
\label{sec:setup}

\textbf{Base student and teacher models.} Following~\citep{zhou2026dllm,cheng2025sdar}, we use the Qwen3 family (0.6B to 8B) as our base ARLMs for conversion to DLMs~\citep{qwen3}. Following our self-distillation setup, the teacher and student are same models, i.e., 4B ARLM teaches 4B student, differing only in use of logit shifting for the ARLM and their attention mask, i.e., the teacher has causal attention and the student uses block-wise causal attention.

\textbf{Dataset Preparation.} %\dk{Give more details on how we create the training data. }  
% We train both general-purpose and task-specialized versions of OPDLM. 
% To address a notable gap in current DLM literature, the lack of transparency in pretraining data,  \dk{The focus should be on explaining how we create the data set. Since the submission is anonymous, you can say that ``our curated data set is available in the anonymous repo [link]''} \sout{we openly release our curated dataset}
We train OPDLM on a curated corpus of $\sim$60K samples spanning four domains: math (20,222 samples), code (21,594 samples), science (10,000 samples) and chat (10,000 samples), with details in~\cref{sec:dataset}. Since OPDLM is an on-policy method, we only keep the prompts of those datasets.

For evaluation, we report results across different categories: general knowledge, math, and coding. Unless otherwise specified, all benchmarks are evaluated under greedy static decoding~\citep{nie2025large} with block size=4. Detailed hyperparameters can be found in~\cref{app:hyperparameter}.

\paragraph{Baselines.} We compare OPDLM against four representative DLMs: \textbf{LLaDA}~\citep{nie2025large}, trained from scratch with a full-attention diffusion objective; \textbf{Dream}~\citep{ye2025dream7bdiffusionlarge}, also using a full-attention diffusion objective but initialized from a pretrained AR checkpoint; and \textbf{SDAR}~\citep{cheng2025sdar} and \textbf{Fast-dLLM-v2}~\citep{wu2025fast}, which adapt pretrained AR models into block diffusion models via off-policy conversion. Unless otherwise stated, all reported results use static decoding (one token per step).

\subsection{OPDLM as a General-Purpose DLM}
\label{sec:main_results}

\begin{table}[t]
\centering
\small
\setlength{\tabcolsep}{4pt}
\caption{Zero-shot extended thinking in OPDLM. Although not explicitly trained to think, OPDLM retains the pre-trained ARLM prior, enabling zero-shot extended thinking at inference.}
\label{tab:extended_think}
\begin{tabular}{lrrrr}
\toprule
& \multicolumn{2}{c}{\textbf{OPDLM-4B}} & \multicolumn{2}{c}{\textbf{OPDLM-8B}} \\
\cmidrule(lr){2-3}\cmidrule(lr){4-5}
\textbf{Benchmark} & \textit{non-think} & \textit{think@eval} & \textit{non-think} & \textit{think@eval} \\
\midrule
GSM8K      & 87.6 & 85.3 & 87.1 & 88.0 \\
MATH-500   & 72.8 & 75.0 & 71.2 & 75.6 \\
AIME-24    & 14.4 & 11.2 & 14.7 & 18.6 \\
AIME-25    & 12.6 & 13.6 & 12.4 & 19.4 \\
LMB-Hard   & 11.1 & 17.8 & 20.0 & 17.8 \\
ZebraLogic & 10.5 & 9.5  & 12.9 & 17.3 \\
\bottomrule
\end{tabular}
\end{table}

\begin{table}[t]
\centering
\small
\setlength{\tabcolsep}{4pt}
\caption{Zero-shot multilingual results at 4B and 8B scales. Despite no multilingual data in training, OPDLM retains substantial multilingual ability; SDAR serves as an oracle reference.}
\label{tab:multilingual}
\begin{tabular}{lcccc}
\toprule
\textbf{Model} & \makecell{\textbf{MMMLU}\\\textbf{-lite}} & \makecell{\textbf{INCLUDE}\\\textbf{-lite}} & \makecell{\textbf{MT-AIME}\\\textbf{2024}} & \makecell{\textbf{MLogiQA}} \\
\midrule
SDAR-4B          & 50.7 & 53.3 & 3.0 & 46.5 \\
\rowcolor{grad1}
OPDLM-4B         & 51.6 & 49.6 & 5.3 & 46.5 \\
\midrule
Fast-dLLM-v2-7B  & 51.5 & 45.1 & 4.3 & 42.6 \\
SDAR-8B          & 60.8 & 57.8 & 4.0 & 46.3 \\
\rowcolor{grad1}
OPDLM-8B         & 56.0 & 51.9 & 7.9 & 42.0 \\
\bottomrule
\end{tabular}
\end{table}

As shown in \cref{tab:main}, OPDLM achieves performance competitive with strong AR-to-diffusion baselines while using two to three orders of magnitude less tokens. 
SDAR remains the strong baseline on several benchmarks; however, this comparison is confounded by two factors: SDAR is trained on $\sim$55B tokens of undisclosed data, whereas OPDLM uses only $\sim$66M-76M tokens from the public corpus. We provide a controlled, like-for-like comparison under matched data and compute in~\cref{sec:sft}, which isolates the contribution of our on-policy distillation objective from data scale and curation effects.

We highlight two additional findings. First, the comparative advantage of OPDLM correlates directly with task complexity. Although baseline models maintain an edge on saturated datasets like GSM8K, OPDLM achieves comparable or superior performance on highly rigorous benchmarks such as GPQA-Diamond, AIME, and LiveCodeBench (note that AIME and GPQA results are averaged across 32 and 8 seeds, respectively, following \citet{cheng2025sdar}). Second, OPDLM effectively preserves the structural priors of its base ARLM. As a result, OPDLM exhibits zero-shot retention of capabilities entirely absent from the training distribution. This includes extended reasoning (\cref{tab:extended_think}), where we prompt the model at evaluation time to produce its reasoning within \texttt{<think></think>} tags before the final answer, and multilingual proficiency (\cref{tab:multilingual}).

\subsection{Ablation Study: On-Policy vs.\ Off-Policy Distillation}
\label{sec:sft}
% \begin{wraptable}{r}{0.57\linewidth}

% \textbf{Comparison of OPDLM and Vanilla pretraining on the same data distribution}. We first compare OPDLM's efficiency against standard ARLM-to-DLM pretraining via the standard DLM loss. For fairness, we match training FLOPs (see Appendix\cref{sec:flop_calc} for calculation details) and compare against supervised pretraining (\textbf{SFT}) on the same data distribution. We also ablate standard knowledge distillation (\textbf{KD})~\citep{Hintonetal2015}. To resolve the off-policy nature of MDLM pretraining, we align the forward and reverse diffusion processes to match the partially masked states encountered during inference, yielding two alternative baselines. First, \textbf{$\text{SFT}_{\text{on}}$} uses an on-policy forward diffusion process but off-policy data and supervision: offline dataset samples are masked using a model-induced trajectory. Second, $\text{OPDLM}_{\text{off}}$ employs an off-policy forward and on-policy reverse setting, applying random masking directly to DLM-generated tokens. These base names reflect the traditional regime where the on/off-policy distinction is strictly defined by the data source: $\text{SFT}_{\text{on}}$ relies on offline data, while $\text{OPDLM}_{\text{off}}$ operates on the model's online generations.

% As can be seen in \cref{tab:ablation_objective}, the OPDLM performs the best when matched, highlighting quicker convergence. \shubham{remaining observation needed after results.}

\textbf{Setup.} In this section, we compare off-policy distillation against OPDLM under matched data: both train on the same prompt corpus for a single data epoch, differing only in the training objective. \textbf{Off-Policy} Distillation~\citep{Hintonetal2015} matches soft teacher distributions on offline teacher responses with random masking. We defer the comparison of SFT and OPDLM on the same data to the appendix.

OPDLM eliminates the training-inference divide of DLMs by training on states generated by the reverse diffusion process of the model itself. To achieve this, OPDLM modifies both the forward diffusion (the masking trajectory) and the reverse diffusion (the target response generation) to be entirely on-policy. To isolate the contribution of the on-policy masking trajectory, we introduce an intermediate baseline, \textbf{OPDLM\textsubscript{off}}, which keeps the reverse diffusion on-policy but reverts the forward diffusion to standard random corruption.
OPDLM\textsubscript{off} ablates the impact of selecting masked states from the decoded trajectory by instead applying standard random corruption to the final generated sequence. We generate a terminal sequence $\vx_0$ using the student's native decoding ($\vx_0 \sim p_\theta$) and compute ARLM soft targets exactly as in OPDLM. To construct the training state, similar to block diffusion, we sample a time step $t \sim \mathcal{U}(0,1)$ and apply independent random masking $\vx_t \sim q^{t\mid0}(\cdot \mid \vx_0)$, yielding the objective:
\begin{equation}
  \mathcal{L}_{\mathrm{OPDLM_{off}}}(\theta) = \mathbb{E}_{\substack{\vx_0 \sim p_\theta, \ t \sim \mathcal{U}(0,1), \ \vx_t \sim q^{t\mid0}(\cdot \mid \vx_0)}} \left[ \sum_{b=1}^{B} w(t) \sum_{i \in \mathcal{M}(\vx_t^b)} D_{\mathrm{KL}} \left( p_{\mathrm{ARLM}}(\cdot \mid \vx_0^{<b}, \vx_0^{b, <i}) \;\middle\|\; p_\theta^{b,i}(\cdot \mid \vx_0^{<b}, \vx_t^b) \right) \right]  
\end{equation}

% To isolate the contribution of OPDLM's two on-policy components, we further introduce two intermediate variants. \textbf{SFT\textsubscript{on}} keeps on-policy masking (mask states are drawn from a student-induced trajectory) but uses offline teacher responses; \textbf{OPDLM\textsubscript{off}} keeps on-policy responses (the student's own rollouts) but applies random masking to them. Full OPDLM combines both on-policy components: on-policy responses and on-policy masking.

\textbf{Results.} \cref{tab:ablation_objective} reports the comparison. Two findings emerge. \textbf{First}, on-policy data is the primary driver of performance. Switching the training data from offline teacher generations to the model's own on-policy generations, moving from the Off-Policy baseline to OPDLM\textsubscript{off} while keeping the ARLM soft targets fixed, yields consistent gains across both the 4B and 8B scales. \textbf{Second}, when using on-policy data, the specific masking trajectory has a limited effect. OPDLM\textsubscript{off} and OPDLM effectively serve as training augmentations of one another. They differ only in whether their masked states are drawn from random corruption or the model's own decoding trajectory, yet both achieve comparable performance across benchmarks. However, we note that this study was done for a block size of 4 and leave a broader study of varying block sizes (beyond the size of 4 used in these experiments) to future work.

% \begin{table}[t]
%     \centering
%     % \small
%     % \vspace{-4.5mm}
%     \setlength{\tabcolsep}{3pt}
%     \caption{Ablation of training objectives under matched data and compute. All variants are trained on the same prompt corpus for one data epoch. \textbf{Bold}: best in row; \underline{underlined}: second best.}
%     \label{tab:ablation_objective}
%     \begin{tabular}{l c c c c}
%         \toprule
%         \textbf{Benchmark} & \textbf{SFT} & \textbf{Off-Policy} & \textbf{OPDLM\textsubscript{off}} & \textbf{OPDLM} \\
%         \midrule
%         MMLU           & 66.6 & 66.3 & 66.7 & {65.5} \\
%         MMLU-Pro       & 48.4 & 47.2 & 47.1 & 46.3 \\
%         GPQA-Diamond   & 29.7 & 30.9 & 29.9 & {29.1} \\
%         IFEval         & 44.7 & 46.2 & 49.0 & \textbf{53.8} \\
%         \midrule
%         GSM8K          & 86.8 & 85.8 & 86.7 & \textbf{87.6} \\
%         MATH500        & 75.4 & 73.2 & 75.8 & {72.8} \\
%         AIME-24        & 12.4 & 10.9 & 11.2 & {14.4} \\
%         AIME-25        & 10.4 & 11.9 & 12.7 & {12.6} \\
%         \midrule
%         HumanEval      & 65.9 & 48.8 & 37.8 & {56.1} \\
%         MBPP           & 58.7 & 54.5 & 56.4 & {57.7} \\
%         LCB-v6         & 18.1 & 6.4  & 11.7 & 10.4 \\
%         Codeforces     & 8.8  & 2.4  & 4.5  & 5.0 \\
%         \bottomrule
%     \end{tabular}
%     % \vspace{-6mm}
% \end{table}

\begin{table}[t]
    \centering
    \small
    \setlength{\tabcolsep}{4pt}
    \caption{Ablation of training objectives under matched data and compute, at 4B and 8B scales. All variants are trained on the same prompt corpus for one data epoch. \textbf{Bold}: best in row within each scale.}
    \label{tab:ablation_objective}
    \begin{tabular}{l ccc ccc}
        \toprule
        & \multicolumn{3}{c}{\textbf{4B}} & \multicolumn{3}{c}{\textbf{8B}} \\
        \cmidrule(lr){2-4}\cmidrule(lr){5-7}
        \textbf{Benchmark} & \textbf{Off-Policy} & \textbf{OPDLM\textsubscript{off}} & \textbf{OPDLM} & \textbf{Off-Policy} & \textbf{OPDLM\textsubscript{off}} & \textbf{OPDLM} \\
        \midrule
        MMLU           & 66.3 & \textbf{66.7} & 65.5 & 65.5 & 69.1 & \textbf{70.9} \\
        MMLU-Pro       & \textbf{47.2} & 47.1 & 46.3 & 51.7 & 52.7 & \textbf{53.7} \\
        GPQA-Diamond   & \textbf{30.9} & 29.9 & 29.1 & 33.6 & \textbf{38.3} & 36.1 \\
        IFEval         & 46.2 & 49.0 & \textbf{53.8} & 43.3 & 48.8 & \textbf{50.1} \\
        \midrule
        GSM8K          & 85.8 & 86.7 & \textbf{87.6} & 85.7 & \textbf{88.1} & 87.1 \\
        MATH500        & 73.2 & \textbf{75.8} & 72.8 & 69.2 & \textbf{73.8} & 71.2 \\
        AIME-24        & 10.9 & 11.2 & \textbf{14.4} & \textbf{17.0} & 12.9 & 14.7 \\
        AIME-25        & 11.9 & \textbf{12.7} & 12.6 & 14.3 & \textbf{15.4} & 12.4 \\
        \midrule
        HumanEval      & 48.8 & 37.8 & \textbf{56.1} & 54.9 & 54.9 & \textbf{59.8} \\
        MBPP           & 54.5 & 56.4 & \textbf{57.7} & \textbf{59.5} & 57.9 & 48.7 \\
        LCB-v6         & 6.4 & \textbf{11.7} & 10.4 & 6.4 & \textbf{11.9} & 9.7 \\
        Codeforces     & 2.4 & 4.5 & \textbf{5.0} & 3.7 & \textbf{4.5} & 3.5 \\
        \bottomrule
    \end{tabular}
\end{table}

\begin{figure}[t]
    \centering

    \begin{subfigure}[t]{0.48\linewidth}
        \centering
        \includegraphics[width=0.95\linewidth]{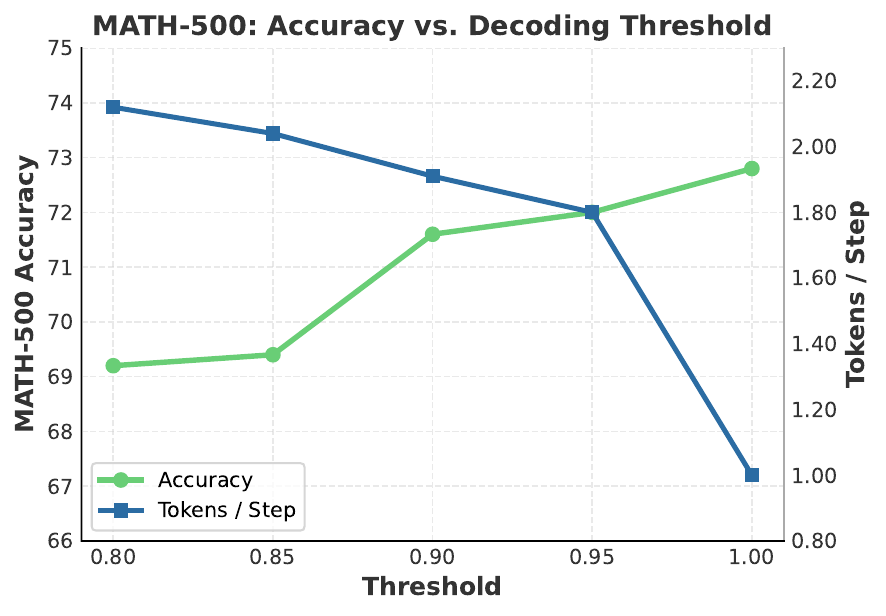}
        % \caption{Effect of decoding threshold.}
        \label{fig:math500_threshold_block4}
    \end{subfigure}
    \hfill
    \begin{subfigure}[t]{0.48\linewidth}
        \centering
        \includegraphics[width=0.95\linewidth]{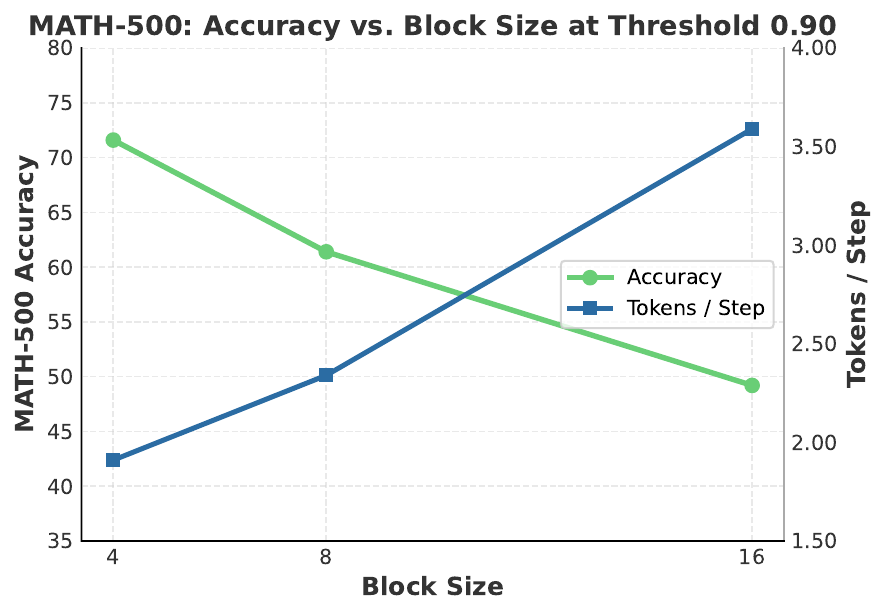}
        % \caption{Effect of block size.}
        \label{fig:math500_blocksize_thr090}
    \end{subfigure}
    \caption{MATH-500 accuracy and average tokens per denoising step under different decoding configurations. \textbf{(a)} OPDLM with block size 4: decreasing the decoding threshold $\gamma$ increases tokens/step, with MATH-500 accuracy as the trade-off. \textbf{(b)} At fixed threshold $\gamma=0.9$: larger training block sizes yield more tokens/step, again at the cost of accuracy.}
    \label{fig:math500_decoding_tradeoffs}
    \vspace{-6mm}
\end{figure}

\begin{table}[H]
% \vspace{-0.4cm}
    \centering
    % \vspace{-\intextsep}
    \caption{OPDLM-MATH compared to TraDo~\citep{wang2025revolutionizing} on math benchmarks. Even without a verifier signal or a pre-training, OPDLM-MATH achieves strong performance, particularly on the harder benchmarks. The "Thinking" variants are trained from scratch as separate models to enable extended reasoning. We also provide a reference comparison to standard DLMs.}
    \label{tab:post_train}
    % \resizebox{\linewidth}{!}{%
    \begin{tabular}{l c c c}
        \toprule
        \textbf{Model} & \textbf{GSM8K} & \textbf{MATH500} & \textbf{AIME24} \\
        \midrule
        \multicolumn{4}{l}{\textit{Reference}} \\
        SDAR-4B-Chat            & 90.2          & 70.2          & 5.0 \\
        LLaDA-8B-Instruct       & 82.5          & 37.3          & 0.5 \\
        Dream-7B-Instruct       & 72.7          & 38.7          & 0.0 \\
        SDAR-8B-Chat            & 91.1          & 74.3          & 11.8 \\
        \midrule
        \multicolumn{4}{l}{\textit{4B scale}} \\
        TraDo-4B-Instruct       & \textbf{91.2} & 75.6          & 8.3 \\
        \rowcolor{grad1}
        OPDLM-MATH-4B           & 83.8          & \textbf{75.8} & \textbf{10.0} \\
        \rowcolor{grad1}
        OPDLM-MATH-4B-Thinking  & \textbf{91.7} & \textbf{90.2} & \textbf{43.3} \\
        \midrule
        \multicolumn{4}{l}{\textit{8B scale}} \\
        TraDo-8B-Instruct       & \textbf{92.3} & \textbf{78.5} & 13.3 \\
        \rowcolor{grad1}
        OPDLM-MATH-8B           & 86.2          & 76.6          & \textbf{23.3} \\
        TraDo-8B-Thinking       & \textbf{94.2} & 87.4          & 35.5 \\
        \rowcolor{grad1}
        OPDLM-MATH-8B-Thinking  & 93.8          & \textbf{92.4} & \textbf{50.0} \\
        \bottomrule
    \end{tabular}%
    % }
    % \vspace{-5mm}
\end{table}

\subsection{Inference Efficiency: Multi-Token Decoding}
\label{sec:inference}

In this section, we briefly explore how OPDLM's inference throughput can be controlled through two complementary mechanisms, namely, the decoding confidence threshold and the block size used in training. More results are provided in~\cref{app:multi_token}.
% Full ablations across thresholds, block sizes, and benchmarks are deferred to Appendix~\ref{app:multi_token}.

\textbf{Effect of decoding threshold.} For a fixed block size, OPDLM admits multiple tokens per denoising step whenever their predicted confidence exceeds a threshold $\gamma$. \cref{fig:math500_threshold_block4} reports this trade-off on MATH-500 with block size 4: throughput increases from 1 token/step at $\gamma=1$ (static decoding) to over 2 tokens/step at $\gamma=0.8$, with a moderate accuracy cost.

\textbf{Effect of block size.} The block size used in training OPDLM models sets an upper bound on the parallelism achievable at inference. \cref{fig:math500_blocksize_thr090} shows that at a fixed threshold $\gamma=0.9$, increasing block size from 4 to 16 raises throughput from $\sim$2 to $\sim$3.5 tokens/step, at some cost of accuracy. 
% The same pattern is visible during training (Figure~\ref{fig:block_size_avg_tok_training}): larger block sizes consistently yield higher average tokens-per-step.

\subsection{OPDLM as a Task-Specific DLM}
\label{sec:posttrain}

% \begin{wraptable}{r}{0.6\linewidth}

Post-training DLMs via reinforcement learning with verifiable rewards (RLVR)~\citep{guo2025deepseek} is a prominent approach for developing task-specialized experts~\citep{wang2025revolutionizing}. However, because our framework distills knowledge directly from an ARLM teacher, we can bypass the intermediate requirement of building a general-purpose DLM. If an expert model is the primary objective, we can achieve this efficiently by training an OPDLM strictly on task-specific data. 

To ensure a fair comparison in~\cref{tab:post_train}, we train OPDLM on the same data used by TraDo~\citep{wang2025revolutionizing}: 8K level 3-5 hard problems from the MATH training set~\citep{hendrycks2021measuring}. The key difference lies in the starting point: TraDo initializes from an SDAR-pretrained DLM and applies RL with verifiable rewards, whereas OPDLM starts directly from the AR Qwen3 checkpoint and applies on-policy distillation with no rewards or pretrained DLM stage. We train two variants: \textbf{OPDLM-MATH}, which distills the teacher in non-thinking mode, and \textbf{OPDLM-MATH-Thinking}, which distills the teacher with its thinking behavior enabled.
\vspace{-2mm}
\section{Conclusion}
\label{sec:conclusion}
\vspace{-2mm}
We introduce On-Policy Diffusion Language Models (\methodname{}), for converting autoregressive language models (ARLMs) into diffusion
language models (DLMs). Existing ARLM-to-DLM conversion methods face two
challenges: retaining knowledge from the original ARLM after changing the
training objective and attention pattern, and reducing the training--inference
state mismatch between forward random masking during training and reverse unmasking at
inference. \methodname{} addresses these challenges by adapting on-policy
distillation to DLM conversion. The student DLM is initialized from the ARLM,
samples its own reverse diffusion trajectories, and is supervised on those
trajectories using token-level distributions from the original frozen ARLM
teacher. Empirically, \methodname{} requires \textbf{15$\times$} to
\textbf{7{,}000$\times$} fewer training tokens than prior DLM baselines while
achieving competitive performance across a broad range of tasks. These results
reframe ARLM-to-DLM conversion as an efficient post-training procedure, opening
a practical path for adapting future ARLMs into diffusion-based language
models.

\section*{Acknowledgments}
This work was supported in part by the National Institutes of Health (NIH) under grant U01AG070112 and by the Advanced Research Projects Agency for Health (ARPA-H) under grant 1AY1AX000053.

% \bibliographystyle{plain}
% \bibliography{whole} 

\clearpage
\bibliographystyle{refstyle}
\bibliography{whole}

@article{guo2025deepseek,
  title={Deepseek-r1: Incentivizing reasoning capability in llms via reinforcement learning},
  author={Guo, Daya and Yang, Dejian and Zhang, Haowei and Song, Junxiao and Zhang, Ruoyu and Xu, Runxin and Zhu, Qihao and Ma, Shirong and Wang, Peiyi and Bi, Xiao and others},
  journal={arXiv preprint arXiv:2501.12948},
  year={2025}
}

@inproceedings{bengio2009curriculum,
  title     = {Curriculum Learning},
  author    = {Bengio, Yoshua and Louradour, J{\'e}r{\^o}me and Collobert, Ronan and Weston, Jason},
  booktitle = {Proceedings of the 26th Annual International Conference on Machine Learning},
  pages     = {41--48},
  year      = {2009},
  publisher = {ACM}
}

@article{yang2025qwen3,
  title={Qwen3 technical report},
  author={Yang, An and Li, Anfeng and Yang, Baosong and Zhang, Beichen and Hui, Binyuan and Zheng, Bo and Yu, Bowen and Gao, Chang and Huang, Chengen and Lv, Chenxu and others},
  journal={arXiv preprint arXiv:2505.09388},
  year={2025}
}

@inproceedings{son2025linguistic,
  title={Linguistic generalizability of test-time scaling in mathematical reasoning},
  author={Son, Guijin and Hong, Jiwoo and Ko, Hyunwoo and Thorne, James},
  booktitle={Proceedings of the 63rd Annual Meeting of the Association for Computational Linguistics (Volume 1: Long Papers)},
  pages={14333--14368},
  year={2025}
}

@inproceedings{zhang2025p,
  title={P-mmeval: A parallel multilingual multitask benchmark for consistent evaluation of llms},
  author={Zhang, Yidan and Wan, Yu and Deng, Boyi and Yang, Baosong and Wei, Hao-Ran and Huang, Fei and Yu, Bowen and Liu, Dayiheng and Lin, Junyang and Zhou, Jingren},
  booktitle={Proceedings of the 2025 Conference on Empirical Methods in Natural Language Processing},
  pages={4809--4836},
  year={2025}
}

@article{romanou2024include,
  title={Include: Evaluating multilingual language understanding with regional knowledge},
  author={Romanou, Angelika and Foroutan, Negar and Sotnikova, Anna and Chen, Zeming and Nelaturu, Sree Harsha and Singh, Shivalika and Maheshwary, Rishabh and Altomare, Micol and Haggag, Mohamed A and Amayuelas, Alfonso and others},
  journal={arXiv preprint arXiv:2411.19799},
  year={2024}
}

@article{hendrycks2021measuring,
  title={Measuring mathematical problem solving with the math dataset},
  author={Hendrycks, Dan and Burns, Collin and Kadavath, Saurav and Arora, Akul and Basart, Steven and Tang, Eric and Song, Dawn and Steinhardt, Jacob},
  journal={arXiv preprint arXiv:2103.03874},
  year={2021}
}

@article{sahoo2024simple,
  title={Simple and effective masked diffusion language models},
  author={Sahoo, Subham S and Arriola, Marianne and Schiff, Yair and Gokaslan, Aaron and Marroquin, Edgar and Chiu, Justin T and Rush, Alexander and Kuleshov, Volodymyr},
  journal={Advances in Neural Information Processing Systems},
  volume={37},
  pages={130136--130184},
  year={2024}
}

@misc{penedo2025codeforces,
      title={CodeForces}, 
      author={Guilherme Penedo and Anton Lozhkov and Hynek Kydlíček and Loubna Ben Allal and Edward Beeching and Agustín Piqueres Lajarín and Quentin Gallouédec and Nathan Habib and Lewis Tunstall and Leandro von Werra},
      year={2025},
      publisher = {Hugging Face},
      journal = {Hugging Face repository},
      howpublished = {\url{https://huggingface.co/datasets/open-r1/codeforces}}
}

@article{cheng2025sdar,
  title   = {{SDAR}: A Synergistic Diffusion-AutoRegression Paradigm for Scalable Sequence Generation},
  author  = {Cheng, Shuang and Bian, Yihan and Liu, Dawei and Jiang, Yuhua and Liu, Yihao and Zhang, Linfeng and Wang, Wenhai and Guo, Qipeng and Chen, Kai and Qi, Biqing and Zhou, Bowen},
  journal = {arXiv preprint arXiv:2510.06303},
  year    = {2025},
  eprint  = {2510.06303},
  archivePrefix = {arXiv},
  primaryClass  = {cs.LG},
  url     = {https://arxiv.org/abs/2510.06303}
}

@article{wang2025revolutionizing,
  title={Revolutionizing reinforcement learning framework for diffusion large language models},
  author={Wang, Yinjie and Yang, Ling and Li, Bowen and Tian, Ye and Shen, Ke and Wang, Mengdi},
  journal={arXiv preprint arXiv:2509.06949},
  year={2025}
}

@article{ho2020denoising,
  title={Denoising diffusion probabilistic models},
  author={Ho, Jonathan and Jain, Ajay and Abbeel, Pieter},
  journal={Advances in neural information processing systems},
  volume={33},
  pages={6840--6851},
  year={2020}
}

@article{hubotter2026sdpo,
  title={Reinforcement Learning via Self-Distillation},
  author={H{\"u}botter, Jonas and L{\"u}beck, Frederike and Behric, Lejs and Baumann, Anton and Bagatella, Marco and Marta, Daniel and Hakimi, Ido and Shenfeld, Idan and Kleine Buening, Thomas and Guestrin, Carlos and Krause, Andreas},
  journal={arXiv preprint arXiv:2601.20802},
  year={2026}
}

@article{shenfeld2026sdft,
  title   = {Self-Distillation Enables Continual Learning},
  author  = {Shenfeld, Idan and Damani, Mehul and H{\"u}botter, Jonas and Agrawal, Pulkit},
  journal = {arXiv preprint arXiv:2601.19897},
  year    = {2026},
  eprint  = {2601.19897},
  archivePrefix = {arXiv},
  primaryClass  = {cs.LG},
  url     = {https://arxiv.org/abs/2601.19897}
}

@article{zhao2026self,
  title={Self-Distilled Reasoner: On-Policy Self-Distillation for Large Language Models},
  author={Zhao, Siyan and Xie, Zhihui and Liu, Mengchen and Huang, Jing and Pang, Guan and Chen, Feiyu and Grover, Aditya},
  journal={arXiv preprint arXiv:2601.18734},
  year={2026}
}

@article{mehta2026breaking,
  title={Breaking the Limits of Open-Weight CLIP: An Optimization Framework for Self-supervised Fine-tuning of CLIP},
  author={Mehta, Anant and Wei, Xiyuan and Chen, Xingyu and Yang, Tianbao},
  journal={arXiv preprint arXiv:2601.09859},
  year={2026}
}

@article{song2019generative,
  title={Generative modeling by estimating gradients of the data distribution},
  author={Song, Yang and Ermon, Stefano},
  journal={Advances in neural information processing systems},
  volume={32},
  year={2019}
}

@article{qwen3,
  title   = {Qwen3 Technical Report},
  author  = {{Qwen Team}},
  journal = {arXiv preprint arXiv:2505.09388},
  year    = {2025},
  url     = {https://arxiv.org/abs/2505.09388}
}

@inproceedings{dagger,
  title={A reduction of imitation learning and structured prediction to no-regret online learning},
  author={Ross, St{\'e}phane and Gordon, Geoffrey and Bagnell, Drew},
  booktitle={Proceedings of the fourteenth international conference on artificial intelligence and statistics},
  pages={627--635},
  year={2011},
  organization={JMLR Workshop and Conference Proceedings}
}

@techreport{deepseekv4,
  title       = {DeepSeek-V4 Technical Report},
  author      = {{DeepSeek-AI}},
  institution = {DeepSeek-AI},
  year        = {2026},
  url         = {https://huggingface.co/deepseek-ai/DeepSeek-V4-Pro/blob/main/DeepSeek_V4.pdf}
}

@article{ye2025dream7bdiffusionlarge,
  title   = {Dream 7B: Diffusion Large Language Models},
  author  = {Ye, Jiacheng and Xie, Zhihui and Zheng, Lin and Gao, Jiahui and Wu, Zirui and Jiang, Xin and Li, Zhenguo and Kong, Lingpeng},
  journal = {arXiv preprint arXiv:2508.15487},
  year    = {2025},
  eprint  = {2508.15487},
  archivePrefix = {arXiv},
  primaryClass  = {cs.CL},
  url     = {https://arxiv.org/abs/2508.15487}
}

@article{Nietal2025,
  author  = {Ni, Jinjie and Liu, Qian and Dou, Longxu and Du, Chao and Wang, Zili and Yan, Hang and Pang, Tianyu and Shieh, Michael Qizhe},
  title   = {Diffusion Language Models are Super Data Learners},
  journal = {arXiv},
  year    = {2025},
  doi     = {10.48550/arxiv.2511.03276}
}

@article{Gaoetal2025,
  author  = {Gao, Zitian and Luo, Haoming and Chen, Lynx and Liu, Jason Klein and Tao, Ran and Zhou, Joey and Dai, Bryan},
  title   = {What Makes Diffusion Language Models Super Data Learners?},
  journal = {arXiv},
  year    = {2025},
  doi     = {10.48550/arxiv.2510.04071}
}

@inproceedings{
parashar2026curriculum,
title={Curriculum Reinforcement Learning from Easy to Hard Tasks Improves {LLM} Reasoning},
author={Shubham Parashar and Shurui Gui and Xiner Li and Hongyi Ling and Sushil Vemuri and Blake Olson and Eric Li and Yu Zhang and James Caverlee and Dileep Kalathil and Shuiwang Ji},
booktitle={The Fourteenth International Conference on Learning Representations},
year={2026},
url={https://openreview.net/forum?id=KJvHnl3kUv}
}

@article{Hintonetal2015,
  author  = {Hinton, Geoffrey and Vinyals, Oriol and Dean, Jeff},
  title   = {Distilling the Knowledge in a Neural Network},
  journal = {arXiv},
  year    = {2015},
  doi     = {10.48550/arxiv.1503.02531}
}

@article{Xieetal2024,
  author  = {Xie, Yong and Aggarwal, Karan and Ahmad, Aitzaz},
  title   = {Efficient Continual Pre-training for Building Domain Specific Large Language Models},
  journal = {Findings of the Association for Computational Linguistics ACL 2024},
  year    = {2024},
  pages   = {10184--10201},
  doi     = {10.18653/v1/2024.findings-acl.606}
}

@article{llada2,
  title={Llada2. 0: Scaling up diffusion language models to 100b},
  author={Bie, Tiwei and Cao, Maosong and Chen, Kun and Du, Lun and Gong, Mingliang and Gong, Zhuochen and Gu, Yanmei and Hu, Jiaqi and Huang, Zenan and Lan, Zhenzhong and others},
  journal={arXiv preprint arXiv:2512.15745},
  year={2025}
}

@article{nbdiff,
  title   = {From Next-Token to Next-Block: A Principled Adaptation Path for Diffusion LLMs},
  author  = {Tian, Yuchuan and Liang, Yuchen and Sun, Jiacheng and Zhang, Shuo and Yang, Guangwen and Shu, Yingte and Fang, Sibo and Guo, Tianyu and Han, Kai and Xu, Chao and Chen, Hanting and Chen, Xinghao and Wang, Yunhe},
  journal = {arXiv preprint arXiv:2512.06776},
  year    = {2026},
  eprint  = {2512.06776},
  archivePrefix = {arXiv},
  primaryClass  = {cs.CL},
  url     = {https://arxiv.org/abs/2512.06776}
}

@article{efficientdlm,
  title   = {Efficient-DLM: From Autoregressive to Diffusion Language Models, and Beyond in Speed},
  author  = {Fu, Yonggan and Whalen, Lexington and Ye, Zhifan and Dong, Xin and Diao, Shizhe and Liu, Jingyu and Wu, Chengyue and Zhang, Hao and Xie, Enze and Han, Song and Khadkevich, Maksim and Kautz, Jan and Lin, Yingyan Celine and Molchanov, Pavlo},
  journal = {arXiv preprint arXiv:2512.14067},
  year    = {2026},
  eprint  = {2512.14067},
  archivePrefix = {arXiv},
  primaryClass  = {cs.CL},
  url     = {https://arxiv.org/abs/2512.14067}
}

@article{hendrycks2020measuring,
  title={Measuring massive multitask language understanding},
  author={Hendrycks, Dan and Burns, Collin and Basart, Steven and Zou, Andy and Mazeika, Mantas and Song, Dawn and Steinhardt, Jacob},
  journal={arXiv preprint arXiv:2009.03300},
  year={2020}
}

@inproceedings{wang2024mmluprorobustchallengingmultitask,
  title     = {{MMLU-Pro}: A More Robust and Challenging Multi-Task Language Understanding Benchmark},
  author    = {Wang, Yubo and Ma, Xueguang and Zhang, Ge and Ni, Yuansheng and Chandra, Abhranil and Guo, Shiguang and Ren, Weiming and Arulraj, Aaran and He, Xuan and Jiang, Ziyan and Li, Tianle and Ku, Max and Wang, Kai and Zhuang, Alex and Fan, Rongqi and Yue, Xiang and Chen, Wenhu},
  booktitle = {Advances in Neural Information Processing Systems},
  volume    = {37},
  year      = {2024},
  eprint    = {2406.01574},
  archivePrefix = {arXiv},
  primaryClass  = {cs.CL},
  url       = {https://proceedings.neurips.cc/paper_files/paper/2024/hash/ad236edc564f3e3156e1b2feafb99a24-Abstract.html}
}

@article{austin2021structured,
  title={Structured denoising diffusion models in discrete state-spaces},
  author={Austin, Jacob and Johnson, Daniel D and Ho, Jonathan and Tarlow, Daniel and Van Den Berg, Rianne},
  journal={Advances in neural information processing systems},
  volume={34},
  pages={17981--17993},
  year={2021}
}

@inproceedings{
nie2025large,
title={Large Language Diffusion Models},
author={Shen Nie and Fengqi Zhu and Zebin You and Xiaolu Zhang and Jingyang Ou and Jun Hu and JUN ZHOU and Yankai Lin and Ji-Rong Wen and Chongxuan Li},
booktitle={The Thirty-ninth Annual Conference on Neural Information Processing Systems},
year={2025},
url={https://openreview.net/forum?id=KnqiC0znVF}
}

@inproceedings{
arriola2025block,
title={Block Diffusion: Interpolating Between Autoregressive and Diffusion Language Models},
author={Marianne Arriola and Aaron Gokaslan and Justin T Chiu and Zhihan Yang and Zhixuan Qi and Jiaqi Han and Subham Sekhar Sahoo and Volodymyr Kuleshov},
booktitle={The Thirteenth International Conference on Learning Representations},
year={2025},
url={https://arxiv.org/abs/2503.09573}
}

@inproceedings{
agarwal2024onpolicy,
title={On-Policy Distillation of Language Models: Learning from Self-Generated Mistakes},
author={Rishabh Agarwal and Nino Vieillard and Yongchao Zhou and Piotr Stanczyk and Sabela Ramos Garea and Matthieu Geist and Olivier Bachem},
booktitle={The Twelfth International Conference on Learning Representations},
year={2024},
url={https://openreview.net/forum?id=3zKtaqxLhW}
}

@article{dhariwal2021diffusion,
  title={Diffusion models beat gans on image synthesis},
  author={Dhariwal, Prafulla and Nichol, Alexander},
  journal={Advances in neural information processing systems},
  volume={34},
  pages={8780--8794},
  year={2021}
}

@article{zhou2024simple,
  title={Simple and Fast Distillation of Diffusion Models},
  author={Zhou, Zhenyu and Chen, Defang and Wang, Can and Chen, Chun and Lyu, Siwei},
  journal={arXiv preprint arXiv:2409.19681},
  year={2024}
}

@article{ho2022video,
  title={Video diffusion models},
  author={Ho, Jonathan and Salimans, Tim and Gritsenko, Alexey and Chan, William and Norouzi, Mohammad and Fleet, David J},
  journal={Advances in Neural Information Processing Systems},
  volume={35},
  pages={8633--8646},
  year={2022}
}

@article{pearce2023imitating,
  title={Imitating human behaviour with diffusion models},
  author={Pearce, Tim and Rashid, Tabish and Kanervisto, Anssi and Bignell, Dave and Sun, Mingfei and Georgescu, Raluca and Macua, Sergio Valcarcel and Tan, Shan Zheng and Momennejad, Ida and Hofmann, Katja and others},
  journal={arXiv preprint arXiv:2301.10677},
  year={2023}
}

@article{hayes2024simulating,
  title={Simulating 500 million years of evolution with a language model},
  author={Hayes, Tomas and Rao, Roshan and Akin, Halil and Sofroniew, Nicholas J and Oktay, Deniz and Lin, Zeming and Verkuil, Robert and Tran, Vincent Q and Deaton, Jonathan and Wiggert, Marius and others},
  journal={bioRxiv},
  pages={2024--07},
  year={2024},
  publisher={Cold Spring Harbor Laboratory}
}

@inproceedings{
yu2026dapo,
title={{DAPO}: An Open-Source {LLM} Reinforcement Learning System at Scale},
author={Qiying Yu and Zheng Zhang and Ruofei Zhu and Yufeng Yuan and Xiaochen Zuo and YuYue and Weinan Dai and Tiantian Fan and Gaohong Liu and Juncai Liu and LingJun Liu and Xin Liu and Haibin Lin and Zhiqi Lin and Bole Ma and Guangming Sheng and Yuxuan Tong and Chi Zhang and Mofan Zhang and Ru Zhang and Wang Zhang and Hang Zhu and Jinhua Zhu and Jiaze Chen and Jiangjie Chen and Chengyi Wang and Hongli Yu and Yuxuan Song and Xiangpeng Wei and Hao Zhou and Jingjing Liu and Wei-Ying Ma and Ya-Qin Zhang and Lin Yan and Yonghui Wu and Mingxuan Wang},
booktitle={The Thirty-ninth Annual Conference on Neural Information Processing Systems},
year={2026},
url={https://openreview.net/forum?id=2a36EMSSTp}
}

@article{du2025nemotronmathefficientlongcontextdistillation,
  title   = {Nemotron-Math: Efficient Long-Context Distillation of Mathematical Reasoning from Multi-Mode Supervision},
  author  = {Du, Wei and Toshniwal, Shubham and Kisacanin, Branislav and Mahdavi, Sadegh and Moshkov, Ivan and Armstrong, George and Ge, Stephen and Minasyan, Edgar and Chen, Feng and Gitman, Igor},
  journal = {arXiv preprint arXiv:2512.15489},
  year    = {2025},
  eprint  = {2512.15489},
  archivePrefix = {arXiv},
  primaryClass  = {cs.AI},
  url     = {https://arxiv.org/abs/2512.15489}
}

@article{li2023tacotopicsalgorithmiccode,
  title   = {{TACO}: Topics in Algorithmic Code Generation Dataset},
  author  = {Li, Rongao and Fu, Jie and Zhang, Bo-Wen and Huang, Tao and Sun, Zhihong and Lyu, Chen and Liu, Guang and Jin, Zhi and Li, Ge},
  journal = {arXiv preprint arXiv:2312.14852},
  year    = {2023},
  eprint  = {2312.14852},
  archivePrefix = {arXiv},
  primaryClass  = {cs.AI},
  url     = {https://arxiv.org/abs/2312.14852}
}

@inproceedings{xu2025kodcodediversechallengingverifiable,
  title     = {{KodCode}: A Diverse, Challenging, and Verifiable Synthetic Dataset for Coding},
  author    = {Xu, Zhangchen and Liu, Yang and Yin, Yueqin and Zhou, Mingyuan and Poovendran, Radha},
  booktitle = {Findings of the Association for Computational Linguistics: ACL 2025},
  year      = {2025},
  eprint    = {2503.02951},
  archivePrefix = {arXiv},
  primaryClass  = {cs.LG},
  url       = {https://aclanthology.org/2025.findings-acl.365/}
}

@inproceedings{zeng2025acecoderacingcoderrl,
  title     = {{ACECODER}: Acing Coder RL via Automated Test-Case Synthesis},
  author    = {Zeng, Huaye and Jiang, Dongfu and Wang, Haozhe and Nie, Ping and Chen, Xiaotong and Chen, Wenhu},
  booktitle = {Proceedings of the 63rd Annual Meeting of the Association for Computational Linguistics (Volume 1: Long Papers)},
  pages     = {12023--12040},
  year      = {2025},
  eprint    = {2502.01718},
  archivePrefix = {arXiv},
  primaryClass  = {cs.SE},
  url       = {https://aclanthology.org/2025.acl-long.587/}
}

@article{nvidia2025nvidianemotronnano2,
  title   = {{NVIDIA} Nemotron Nano 2: An Accurate and Efficient Hybrid Mamba-Transformer Reasoning Model},
  author  = {{NVIDIA} and Basant, Aarti and Khairnar, Abhijit and Paithankar, Abhijit and Khattar, Abhinav and Renduchintala, Adithya and Malte, Aditya and Bercovich, Akhiad and Hazare, Akshay and Rico, Alejandra and Ficek, Aleksander and Kondratenko, Alex and Shaposhnikov, Alex and Bukharin, Alexander and others},
  journal = {arXiv preprint arXiv:2508.14444},
  year    = {2025},
  eprint  = {2508.14444},
  archivePrefix = {arXiv},
  primaryClass  = {cs.CL},
  url     = {https://arxiv.org/abs/2508.14444}
}

@article{cobbe2021gsm8k,
  title={Training Verifiers to Solve Math Word Problems},
  author={Cobbe, Karl and Kosaraju, Vineet and Bavarian, Mohammad and Chen, Mark and Jun, Heewoo and Kaiser, Lukasz and Plappert, Matthias and Tworek, Jerry and Hilton, Jacob and Nakano, Reiichiro and Hesse, Christopher and Schulman, John},
  journal={arXiv preprint arXiv:2110.14168},
  year={2021}
}

@article{humanevalchen2021codex,
  title={Evaluating Large Language Models Trained on Code},
  author={Mark Chen and Jerry Tworek and Heewoo Jun and Qiming Yuan and Henrique Ponde de Oliveira Pinto and Jared Kaplan and Harri Edwards and Yuri Burda and Nicholas Joseph and Greg Brockman and Alex Ray and Raul Puri and Gretchen Krueger and Michael Petrov and Heidy Khlaaf and Girish Sastry and Pamela Mishkin and Brooke Chan and Scott Gray and Nick Ryder and Mikhail Pavlov and Alethea Power and Lukasz Kaiser and Mohammad Bavarian and Clemens Winter and Philippe Tillet and Felipe Petroski Such and Dave Cummings and Matthias Plappert and Fotios Chantzis and Elizabeth Barnes and Ariel Herbert-Voss and William Hebgen Guss and Alex Nichol and Alex Paino and Nikolas Tezak and Jie Tang and Igor Babuschkin and Suchir Balaji and Shantanu Jain and William Saunders and Christopher Hesse and Andrew N. Carr and Jan Leike and Josh Achiam and Vedant Misra and Evan Morikawa and Alec Radford and Matthew Knight and Miles Brundage and Mira Murati and Katie Mayer and Peter Welinder and Bob McGrew and Dario Amodei and Sam McCandlish and Ilya Sutskever and Wojciech Zaremba},
  year={2021},
  eprint={2107.03374},
  archivePrefix={arXiv},
  primaryClass={cs.LG}
}

@inproceedings{rein2023gpqagraduatelevelgoogleproofqa,
  title     = {{GPQA}: A Graduate-Level Google-Proof Q\&A Benchmark},
  author    = {Rein, David and Hou, Betty Li and Stickland, Asa Cooper and Petty, Jackson and Pang, Richard Yuanzhe and Dirani, Julien and Michael, Julian and Bowman, Samuel R.},
  booktitle = {First Conference on Language Modeling},
  year      = {2024},
  eprint    = {2311.12022},
  archivePrefix = {arXiv},
  primaryClass  = {cs.AI},
  url       = {https://openreview.net/forum?id=Ti67584b98}
}

@inproceedings{jain2024livecodebenchholisticcontaminationfree,
  title     = {{LiveCodeBench}: Holistic and Contamination Free Evaluation of Large Language Models for Code},
  author    = {Jain, Naman and Han, King and Gu, Alex and Li, Wen-Ding and Yan, Fanjia and Zhang, Tianjun and Wang, Sida and Solar-Lezama, Armando and Sen, Koushik and Stoica, Ion},
  booktitle = {The Thirteenth International Conference on Learning Representations},
  year      = {2025},
  eprint    = {2403.07974},
  archivePrefix = {arXiv},
  primaryClass  = {cs.SE},
  url       = {https://openreview.net/forum?id=chfJJYC3iL}
}

@article{zhou2023instructionfollowingevaluationlargelanguage,
  title   = {Instruction-Following Evaluation for Large Language Models},
  author  = {Zhou, Jeffrey and Lu, Tianjian and Mishra, Swaroop and Brahma, Siddhartha and Basu, Sujoy and Luan, Yi and Zhou, Denny and Hou, Le},
  journal = {arXiv preprint arXiv:2311.07911},
  year    = {2023},
  eprint  = {2311.07911},
  archivePrefix = {arXiv},
  primaryClass  = {cs.CL},
  url     = {https://arxiv.org/abs/2311.07911}
}

@inproceedings{cevalhuang2023cevalmultilevelmultidisciplinechinese,
  title     = {{C-Eval}: A Multi-Level Multi-Discipline Chinese Evaluation Suite for Foundation Models},
  author    = {Huang, Yuzhen and Bai, Yuzhuo and Zhu, Zhihao and Zhang, Junlei and Zhang, Jinghan and Su, Tangjun and Liu, Junteng and Lv, Chuancheng and Zhang, Yikai and Lei, Jiayi and Fu, Yao and Sun, Maosong and He, Junxian},
  booktitle = {Advances in Neural Information Processing Systems},
  volume    = {36},
  year      = {2023},
  eprint    = {2305.08322},
  archivePrefix = {arXiv},
  primaryClass  = {cs.CL},
  url       = {https://proceedings.neurips.cc/paper_files/paper/2023/hash/c6ec1844bec96d6d32ae95ae694e23d8-Abstract-Datasets_and_Benchmarks.html}
}

@inproceedings{
white2025livebench,
title={LiveBench: A Challenging, Contamination-Limited {LLM} Benchmark},
author={Colin White and Samuel Dooley and Manley Roberts and Arka Pal and Benjamin Feuer and Siddhartha Jain and Ravid Shwartz-Ziv and Neel Jain and Khalid Saifullah and Sreemanti Dey and Shubh-Agrawal and Sandeep Singh Sandha and Siddartha Venkat Naidu and Chinmay Hegde and Yann LeCun and Tom Goldstein and Willie Neiswanger and Micah Goldblum},
booktitle={The Thirteenth International Conference on Learning Representations},
year={2025},
url={https://openreview.net/forum?id=sKYHBTAxVa}
}

@manual{aime24,
  title  = {American Invitational Mathematics Examination ({AIME}) 2024},
  author = {Zhang, Yifan and {Math-AI Team}},
  year   = {2024},
  url    = {https://huggingface.co/datasets/math-ai/aime24}
}

@article{dekoninck2026matharena,
      title={Beyond Benchmarks: MathArena as an Evaluation Platform for Mathematics with LLMs}, 
      author={Jasper Dekoninck and Nikola Jovanović and Tim Gehrunger and Kári Rögnvalddson and Ivo Petrov and Chenhao Sun and Martin Vechev},
      year={2026},
      eprint={2605.00674},
      archivePrefix={arXiv},
      primaryClass={cs.CL},
      url={https://arxiv.org/abs/2605.00674}, 
}

@inproceedings{fan2024hardmathbenchmarkdatasetchallenging,
  title     = {{HARDMath}: A Benchmark Dataset for Challenging Problems in Applied Mathematics},
  author    = {Fan, Jingxuan and Martinson, Sarah and Wang, Erik Y. and Hausknecht, Kaylie and Brenner, Jonah and Liu, Danxian and Peng, Nianli and Wang, Corey and Brenner, Michael P.},
  booktitle = {The Thirteenth International Conference on Learning Representations},
  year      = {2025},
  eprint    = {2410.09988},
  archivePrefix = {arXiv},
  primaryClass  = {cs.LG},
  url       = {https://openreview.net/forum?id=nDTvP6tBMd}
}

@manual{zebralogic2024,
  title  = {{ZebraLogic}: Benchmarking the Logical Reasoning Ability of Language Models},
  author = {Lin, Bill Yuchen and Le Bras, Ronan and Choi, Yejin},
  year   = {2024},
  url    = {https://huggingface.co/spaces/allenai/ZebraLogic}
}

@article{austin2021programsynthesislargelanguage,
  title   = {Program Synthesis with Large Language Models},
  author  = {Austin, Jacob and Odena, Augustus and Nye, Maxwell and Bosma, Maarten and Michalewski, Henryk and Dohan, David and Jiang, Ellen and Cai, Carrie and Terry, Michael and Le, Quoc and Sutton, Charles},
  journal = {arXiv preprint arXiv:2108.07732},
  year    = {2021},
  eprint  = {2108.07732},
  archivePrefix = {arXiv},
  primaryClass  = {cs.PL},
  url     = {https://arxiv.org/abs/2108.07732}
}

@article{zhou2026dllm,
  title={dllm: Simple diffusion language modeling},
  author={Zhou, Zhanhui and Chen, Lingjie and Tong, Hanghang and Song, Dawn},
  journal={arXiv preprint arXiv:2602.22661},
  year={2026}
}

@article{wu2025fast,
  title={Fast-dllm v2: Efficient block-diffusion llm},
  author={Wu, Chengyue and Zhang, Hao and Xue, Shuchen and Diao, Shizhe and Fu, Yonggan and Liu, Zhijian and Molchanov, Pavlo and Luo, Ping and Han, Song and Xie, Enze},
  journal={arXiv preprint arXiv:2509.26328},
  year={2025}
}

@article{kaplan2020scaling,
  title={Scaling laws for neural language models},
  author={Kaplan, Jared and McCandlish, Sam and Henighan, Tom and Brown, Tom B and Chess, Benjamin and Child, Rewon and Gray, Scott and Radford, Alec and Wu, Jeffrey and Amodei, Dario},
  journal={arXiv preprint arXiv:2001.08361},
  year={2020}
}

%%%%%%%%%%%%%%%%%%%%%%%%%%%%%%%%%%%%%%%%%%%%%%%%%%%%%%%%%%%%

% \newpage
% \input{checklist.tex}

\newpage
\appendix
\phantomsection
\label{sec:appendix}
\addtocontents{toc}{\protect\vspace{1em}}
\addtocontents{toc}{\protect\noindent\textbf{\protect\hyperref[sec:appendix]{Appendix}}\protect\par}
\addtocontents{toc}{\protect\vspace{0.5em}}
% \onecolumn
% \section*{Appendix}
% \newpage
\begin{center}
    {\Large \textbf{Data-Efficient Autoregressive-to-Diffusion Language Models via On-Policy Distillation}}\\[0.5em]
    {\Large Appendix}\\[10mm]
\end{center}

% \section{Technical appendices and supplementary material}
% Technical appendices with additional results, figures, graphs, and proofs may be submitted with the paper submission before the full submission deadline (see above). You can upload a ZIP file for videos or code, but do not upload a separate PDF file for the appendix. There is no page limit for the technical appendices. 

% Note: Think of the appendix as ``optional reading'' for reviewers. The paper must be able to stand alone without the appendix; for example, adding critical experiments that support the main claims to an appendix is inappropriate. 

\section{Additional Experimental Results}
% \shubham{Checklist then Appendix}

\subsection{Results at Smaller Scales}
\label{app:small_scale}

\begin{table*}[ht]
\centering
\footnotesize
\setlength{\tabcolsep}{4pt}
\caption{Main results across 0.6B and 1.7B scales.}
\label{tab:main_small}
\resizebox{\linewidth}{!}{%
\begin{tabular}{l rr rrr}
\toprule
 & \multicolumn{2}{c}{\textbf{0.6B Scale}} & \multicolumn{3}{c}{\textbf{1.7B Scale}} \\
\cmidrule(lr){2-3} \cmidrule(lr){4-6}
\textbf{Benchmark}
  & \textbf{Simple-dLLM} & \textbf{OPDLM-0.6B} & \textbf{SDAR-1.7B}
  & \textbf{Fast-dLLM-v2-1.5B}  & \textbf{OPDLM-1.7B} \\
\midrule
% -------- TOKENS --------
\textit{Tokens} $\downarrow$
  & \cellcolor{grad3}46B
  & {\cellcolor{mygreen}\bb{0.048B}}
  & \cellcolor{grad3}55B
  & \cellcolor{grad4}1B
  & {\cellcolor{mygreen}{\bb{0.072B}}} \\

% -------- FLOPs --------
\textit{FLOPs (1e18)} $\downarrow$
  &  \cellcolor{grad3}165.6
  & {\cellcolor{mygreen}{\bb{0.23}}}
  & \cellcolor{grad3}561.0
  & \cellcolor{grad4}9.0
  & {\cellcolor{mygreen}{\bb{0.98}}} \\
\midrule
MMLU         & 39.1 & \cellcolor{grad1}42.0  & 62.9 & 55.1 & \cellcolor{grad1}53.8 \\
MMLU-Pro     & 13.8 & \cellcolor{grad1}20.4  & 37.0 & 24.1 & \cellcolor{grad1}31.1 \\
GPQA-Diamond & 22.2 & \cellcolor{grad1}24.0  & 29.8 & 24.8 & \cellcolor{grad1}26.8 \\
IFEval       & 35.5 & \cellcolor{grad1}24.8  & 43.4 & 47.0 & \cellcolor{grad1}39.7 \\
\midrule
GSM8K        & 46.3 & \cellcolor{grad1}42.1  & 80.1 & 62.0 & \cellcolor{grad1}71.0 \\
MATH500      & 15.8 & \cellcolor{grad1}28.0  & 63.2 & 39.4 & \cellcolor{grad1}53.0 \\
\midrule
HumanEval-Base & 45.7 & \cellcolor{grad1}20.1 & 59.8  & 43.9 & \cellcolor{grad1}35.4 \\
% HumanEval-Plus & 43.3 & 14.6 & 40.2 & 53.7 & 30.5 \\
MBPP-Base      & 54.0 & \cellcolor{grad1}24.9 & 68.5 & 50.0 & \cellcolor{grad1}45.0 \\
% MBPP-Plus      & 44.2 & 18.3 & 41.3 & 59.8 & 33.3 \\
\bottomrule
\end{tabular}}
\end{table*}

\cref{tab:main_small} reports OPDLM at 0.6B and 1.7B scales against Simple-dLLM~\citep{zhou2026dllm}, Fast-dLLM-v2~\citep{wu2025fast}, and SDAR~\citep{cheng2025sdar}. The trends mirror our main results: OPDLM matches or surpasses comparable-scale baselines on math and reasoning benchmarks while using significantly fewer tokens, and underperforms on coding benchmarks where our data corpus has limited coverage.

\subsection{Effect of Teacher Model Size}
\label{app:teacher_size}

\begin{table}[h]
    \centering
    \caption{Effect of teacher size on OPDLM-MATH-4B and OPDLM-MATH-8B. Same-size teachers (self-distillation) match or outperform a larger Qwen-32B teacher.}
    \label{tab:app_diff_size}
    \begin{tabular}{l c c c}
        \toprule
        \textbf{Model} & \textbf{GSM8K} & \textbf{MATH500} & \textbf{AIME24} \\
        \midrule
        Qwen-4B (Non-thinking)  & 90.2 & 84.8 & 25.0 \\
        Qwen-8B (Non-thinking)  & 91.4 & 87.4 & 29.1 \\
        Qwen-32B (Non-thinking) & 93.6 & 88.6 & 31.0 \\
        \midrule
        OPDLM-MATH-4B (from Qwen-4B)  &  83.8 & 75.8 & 10.0 \\
        OPDLM-MATH-4B (from Qwen-32B) &  84.9 & 72.2 & 10.0 \\
        \midrule
        OPDLM-MATH-8B (from Qwen-8B)  & 86.2 & 76.6 & 23.3 \\
        OPDLM-MATH-8B (from Qwen-32B) & 86.4 & 71.0 & 10.0 \\
        \bottomrule
    \end{tabular}
\end{table}

\begin{table}[t]
\centering
    \caption{Effect of teacher size on OPDLM-0.6B. A larger Qwen3-4B teacher outperforms self-distillation, in contrast to the results in Table~\ref{tab:app_diff_size}.}
\label{tab:teacher_size_06b}
    \begin{tabular}{l c c}
        \toprule
        \textbf{Benchmark} & \textbf{from Qwen3-0.6B} & \textbf{from Qwen3-4B} \\
                           & \textit{(self-distillation)} & \\
        \midrule
        MMLU            & 42.0 & 42.5 \\
        MMLU-Pro        & 20.4 & 19.3 \\
        GPQA-Diamond    & 24.0 & 18.2 \\
        IFEval          & 24.8 & 21.4 \\
        \midrule
        GSM8K           & 42.1 & 47.9 \\
        MATH500         & 28.0 & 36.2 \\
        \midrule
        HumanEval-Base  & 20.1 & 23.8 \\
        MBPP-Base       & 24.9 & 25.4 \\
        \bottomrule
    \end{tabular}
\end{table}

In our main experiments, OPDLM employs self-distillation: a 4B student is distilled from a 4B teacher, and an 8B student from an 8B teacher. A natural question is whether the optimal teacher size depends on the student scale.

\cref{tab:app_diff_size} and \cref{tab:teacher_size_06b} reveal an interesting finding. {For larger students (4B and 8B), self-distillation matches or outperforms a stronger Qwen-32B teacher.} 
{For smaller students (0.6B), self-distillation is no longer sufficient, and a larger Qwen3-4B teacher provides some gains}. 
We hypothesize that a 0.6B teacher does not provide rich enough predictive signal on student-generated rollouts. 
Together, these results suggest the optimal teacher–student configuration depends on student size: same-size teachers suffice for capable students (4B+), while smaller students benefit from moderately larger teachers. A systematic study of this scaling behavior is left to future work.

\subsection{Additional Results for Multi-token Decoding}
\label{app:multi_token}

% \begin{figure}[h]
%     \centering
%     \includegraphics[width=0.9\linewidth]{figs/benchmark_blocksize_dualaxis_thr090.pdf}
%     \caption{MATH-500, GPQA, and MBPP accuracy as a function of average tokens per denoising step and block size for OPDLM with fixed decoding threshold 0.90. As block size increases, benchmark accuracy decreases in exchange for more tokens/step.}
%     \label{fig:gpqa_block4}
% \end{figure}

\begin{figure}[h]
    \centering
    \includegraphics[width=\linewidth]{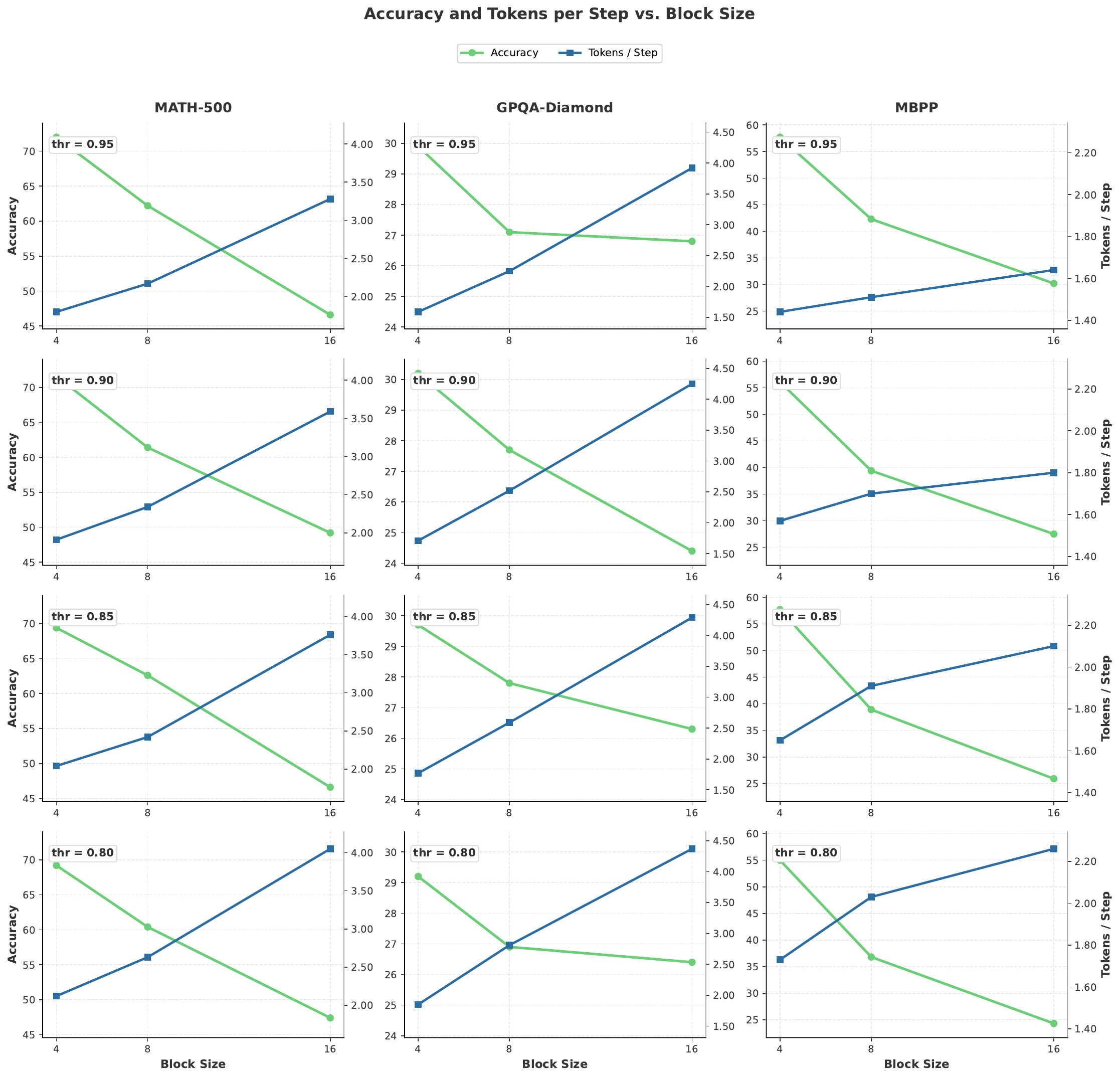}
    \caption{Effect of block size at different decoding thresholds across MATH-500, GPQA-Diamond, and MBPP. Each row corresponds to a fixed threshold $\gamma \in \{0.95, 0.90, 0.85, 0.80\}$.}
    \label{fig:confThresh_block}
\end{figure}

\begin{figure}[h]
    \centering
    \includegraphics[width=\linewidth]{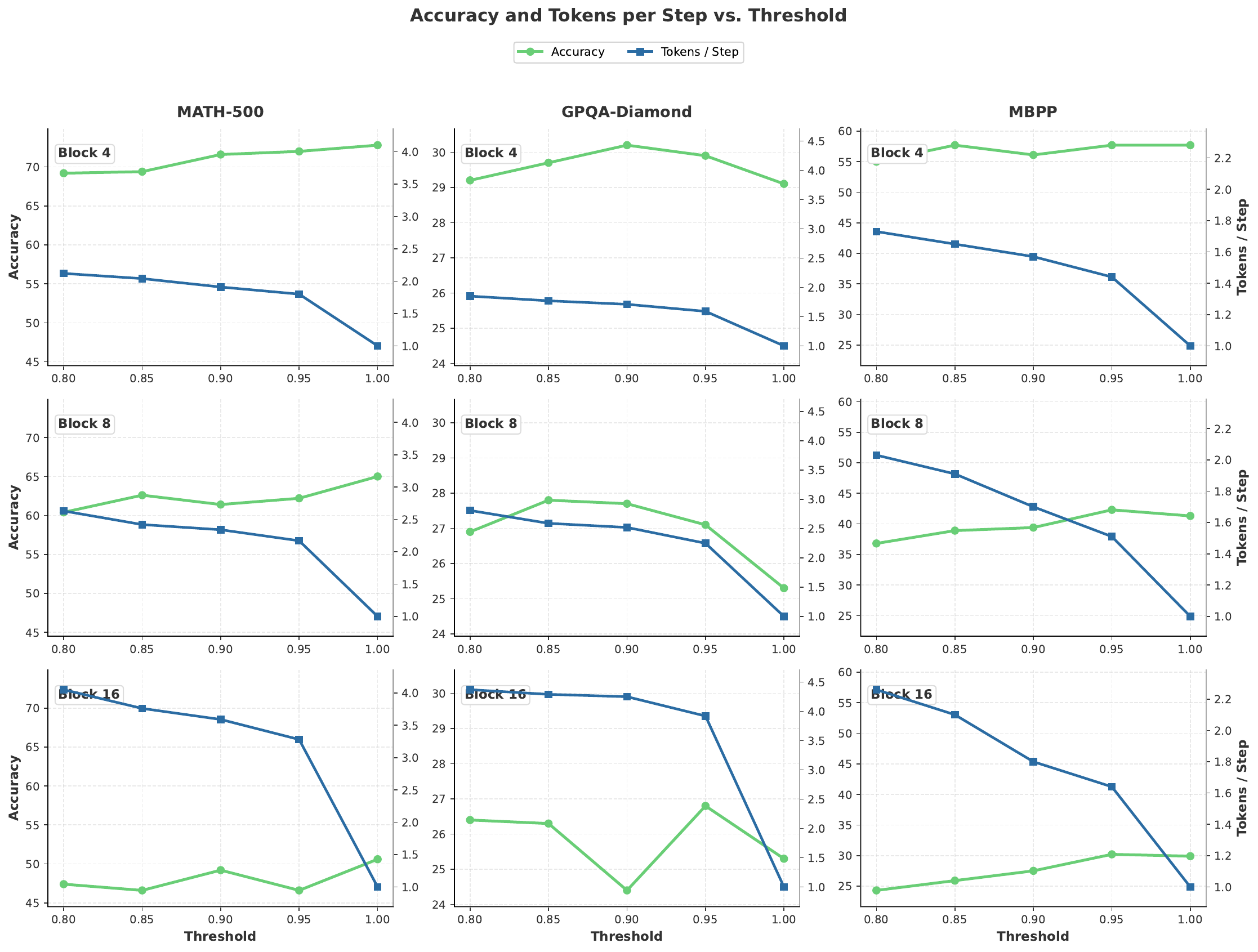}
    \caption{Effect of decoding threshold at different block sizes across MATH-500, GPQA-Diamond, and MBPP. Each row corresponds to a fixed block size $\in \{4, 8, 16\}$.}
    \label{fig:Thresh_grid}
\end{figure}

We extend the analysis in~\cref{sec:inference} along two complementary axes: block size and decoding confidence threshold. \cref{fig:confThresh_block} sweeps block size at four fixed thresholds, while \cref{fig:Thresh_grid} sweeps the threshold at three fixed block sizes. Both figures evaluate three benchmarks (MATH-500, GPQA-Diamond, MBPP) to verify that the trade-off generalizes beyond mathematical reasoning.

Across all configurations, the same accuracy-throughput trade-off holds: larger block sizes and lower thresholds yield more tokens per denoising step at the cost of accuracy. 

\begin{figure}[h]
    \centering
    \includegraphics[width=0.7\linewidth]{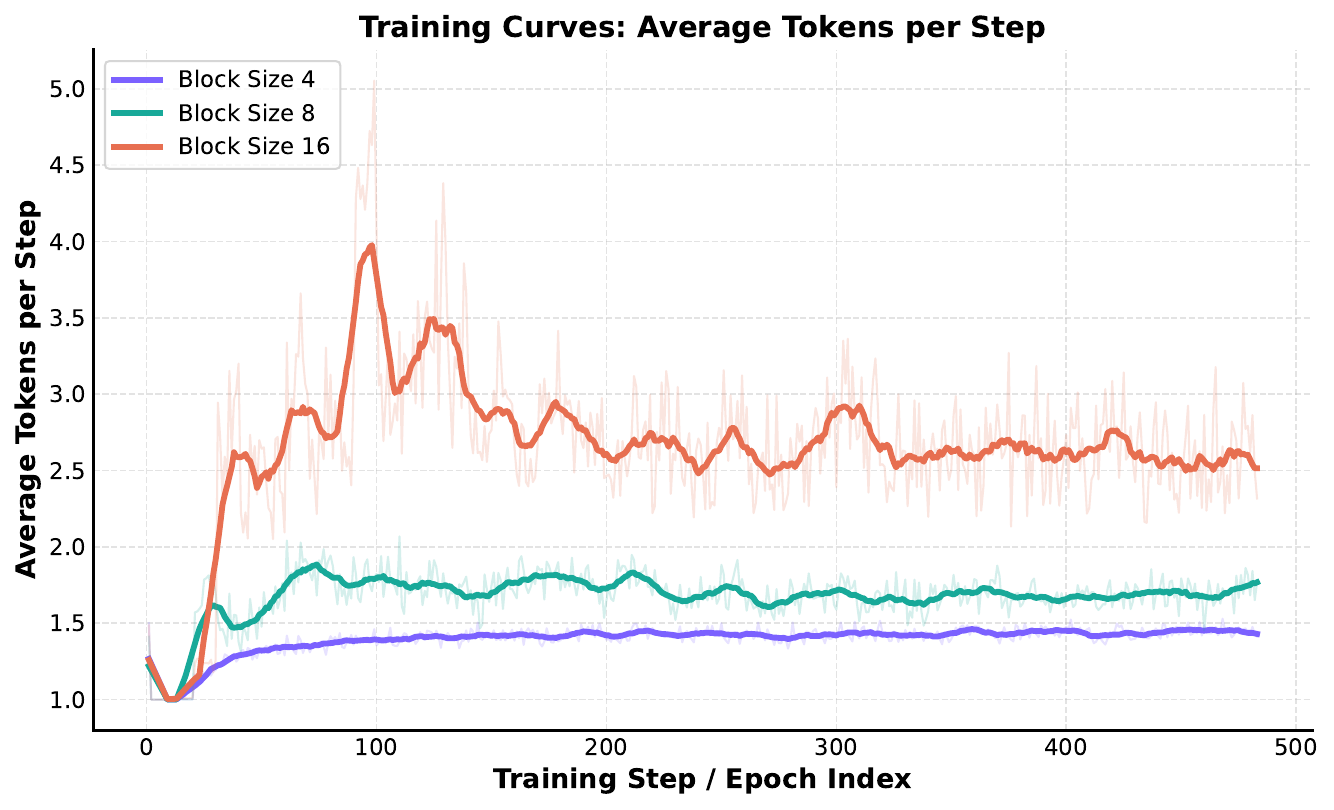}
    \vspace{-8pt}
    \caption{Training curves showing the average number of tokens generated per denoising step across block sizes. Larger block sizes produce more tokens per step, with block size 16 exhibiting the highest average token throughput during training.}
    \label{fig:block_size_avg_tok_training}
\end{figure}

\cref{fig:block_size_avg_tok_training} reports the average number of tokens generated per denoising step over the course of training. Throughput rises rapidly and stabilizes thereafter, suggesting that the model converges quickly to its block-size-determined parallelism level. The relative ordering matches what we observe at inference time: block size 16 achieves the highest throughput throughout training, followed by block size 8 and block size 4, with block size 16 also showing the largest variance during the early phase.

% \begin{figure}[h]
%     \centering
%     \includegraphics[width=0.91\linewidth]{figs/main_figure/benchmark_threshold_grid_3x3.pdf}
%     \caption{3x3 plot}
%     \label{fig:gpqa_block4}
% \end{figure}

\subsection{Multi-Seed Evaluation on GPQA-Diamond, AIME-24, AIME-25}
\label{sec:stat_sig}

\begin{table}[t]
\centering
\small
\setlength{\tabcolsep}{6pt}
\caption{Performance with std on GPQA-Diamond, AIME-24, AIME-25. We evaluate over $N$ random seeds and report mean\,$\pm$\,std.}
\label{tab:small_bench_seeds}
\begin{tabular}{lccc}
\toprule
\textbf{Model} & \textbf{GPQA-Diamond (N=8)} & \textbf{AIME-24 (N=32)} & \textbf{AIME-25 (N=32)} \\
\midrule
OPDLM-0.6B & 24.0\,$\pm$\,2.3 & - & - \\
OPDLM-1.7B & 26.8\,$\pm$\,1.3 & - & - \\
OPDLM-4B   & 29.1\,$\pm$\,2.7 & 14.4\,$\pm$\,3.7 & 12.6\,$\pm$\,3.6 \\
OPDLM-4B think@eval & - & 11.2\,$\pm$\,4.4 & 13.6\,$\pm$\,5.0 \\
OPDLM-8B   & 36.1\,$\pm$\,2.2 & 14.7\,$\pm$\,3.4 & 12.4\,$\pm$\,5.0 \\
OPDLM-8B think@eval & - & 18.6\,$\pm$\,4.2 & 19.4\,$\pm$\,4.2 \\
\bottomrule
\end{tabular}
\end{table}

Table~\ref{tab:small_bench_seeds} reports OPDLM accuracy with standard deviations on GPQA-Diamond, AIME-24, and AIME-25. As these benchmarks are small and single-run scores can fluctuate, we evaluate each model over multiple random seeds and report mean\,$\pm$\,std.

\subsection{Comparison with Supervised Fine-Tuning}

In this sub-section, we extend the off-policy ablation to also measure the impact of training with the BD3LM loss directly on our ARLM-generated data. We generate one response per prompt and apply the BD3LM loss~\citep{arriola2025block}, training for one epoch with block size 4.

\begin{table}[ht]
    \centering
    \small
    \setlength{\tabcolsep}{4pt}
    \caption{Comparison of SFT and OPDLM under matched data and compute at the 4B scale. Both variants are trained on the same prompt corpus for one data epoch. \textbf{Bold}: best in column.}
    \label{tab:sft_vs_opdlm}
    \begin{tabular}{l cccc cccc cc}
        \toprule
        \textbf{Method} & \textbf{MMLU} & \textbf{MMLU-Pro} & \textbf{GPQA} & \textbf{IFEval} & \textbf{GSM8K} & \textbf{MATH500} & \textbf{AIME-24} & \textbf{AIME-25} & \textbf{HumanEval} & \textbf{MBPP} \\
        \midrule
        SFT   & \textbf{66.6} & \textbf{48.4} & \textbf{29.7} & 44.7 & 86.8 & \textbf{75.4} & 12.4 & 10.4 & \textbf{65.9} & \textbf{58.7} \\
        OPDLM & 65.5 & 46.3 & 29.1 & \textbf{53.8} & \textbf{87.6} & 72.8 & \textbf{14.4} & \textbf{12.6} & 56.1 & 57.7 \\
        \bottomrule
    \end{tabular}
\end{table}

\begin{table}[ht]
    \centering
    \small
    \setlength{\tabcolsep}{5pt}
    \caption{Effect of dynamic sampling on OPDLM and SFT at the 4B scale. We report accuracy under static decoding (one token per step) and under dynamic sampling (confidence threshold=$0.9$). \textbf{$\Delta$} is the per-benchmark change under dynamic sampling. We also report the std for AIME benchmarks.}
    \label{tab:dynamic_sampling}
    \begin{tabular}{ll cccc cc}
        \toprule
        \textbf{Method} & \textbf{Decoding} & \textbf{GSM8K} & \textbf{MATH500} & \textbf{AIME-24} & \textbf{AIME-25} & \textbf{HumanEval} & \textbf{MBPP} \\
        \midrule
        \multirow{3}{*}{OPDLM}
          & Static   & 87.6 & 72.8 & 14.4 \tiny$\pm$3.7 & 12.6 \tiny$\pm$3.6 & 53.0 & 58.7 \\
          & Dynamic  & 86.0 & 71.6 & 12.5 \tiny$\pm$3.6 & 14.7 \tiny$\pm$4.2 & 53.0 & 56.1 \\
          & $\Delta$ & $-1.6$ & $-1.2$ & $-1.9$ & $+2.1$ & $0.0$ & $-2.6$ \\
        \midrule
        \multirow{3}{*}{SFT}
          & Static   & 86.8 & 75.4 & 12.4 \tiny$\pm$3.6 & 10.4 \tiny$\pm$3.1 & 65.9 & 59.0 \\
          & Dynamic  & 83.9 & 71.2 & 13.4 \tiny$\pm$4.5 & 9.6 \tiny$\pm$2.2 & 59.8 & 58.5 \\
          & $\Delta$ & $-2.9$ & $-4.2$ & $+1.0$ & $-0.8$ & $-6.1$ & $-0.5$ \\
        \bottomrule
    \end{tabular}%
\end{table}

We first observe that SFT on responses generated by the original ARLM is a simple and efficient recipe, attaining performance comparable to OPDLM across benchmarks at the 4B scale using a block size of 4 (\cref{tab:sft_vs_opdlm}). We view this as a useful and previously underexplored insight; i.e., the setting in which an ARLM produces the exact data later used to convert it into a DLM has rarely been studied, largely because pre-training corpora are seldom released (with the exception of \citet{zhou2024simple}, which instead collects data from various data sources).

Despite their comparable performance on standard benchmarks, SFT and OPDLM diverge significantly under dynamic-sampling-based decoding (\cref{tab:dynamic_sampling}). OPDLM is trained on its own generations, with the data generated using dynamic sampling. As a result, its accuracy under dynamic sampling is highly robust, close to that of one-token-per-step decoding. In contrast, the SFT model is trained on offline responses under random masking and exhibits a much larger average performance drop under dynamic sampling. This drop highlights the train-inference divide native to standard DLMs, a limitation that SFT fails to address. Together, these results suggest that the two approaches occupy complementary regimes: SFT on ARLM responses offers a strong, efficient baseline, while OPDLM provides superior robustness across complex inference-time samplers.

\section{Experiment Details} 
\label{sec:experiment_details}

\subsection{Pretraining Datasets Details}
\label{sec:dataset}

\begin{table}[ht]
    \centering
    \caption{Pretraining dataset details.}
    \label{exp:pretrain}
    \begin{tabular}{l r l}
        \toprule
        \textbf{Domain} & \textbf{\#Samples} & \textbf{Sources} \\
        \midrule
        Math    & 20,222 & DAPO, Nemotron-v2-Math \\
        Code    & 21,594 & TACO, KodCode-Light-RL, AceCode \\
        Science & 10,000 & Nemotron-v2-STEM \\
        Chat    & 10,000 & Nemotron-v2-Chat \\
        \midrule
        \textbf{Total} & \textbf{61,816} & - \\
        \bottomrule
    \end{tabular}
\end{table}

Our pretraining corpus of 62K samples is constructed from a mixture of domains. For mathemtical reasoning, we use 22k samples from DAPO \citep{yu2026dapo} and Nemotron-v2-Math \citep{du2025nemotronmathefficientlongcontextdistillation}. 20k samples of coding data was collated from TACO \citep{li2023tacotopicsalgorithmiccode}, KodCode-Light-RL \citep{xu2025kodcodediversechallengingverifiable}, and AceCode \citep{zeng2025acecoderacingcoderrl}. Finally, from Nemotron-v2 \citep{nvidia2025nvidianemotronnano2} we sample 10k examples of STEM and 10k examples of Chat.

\subsection{Evaluation Benchmark Details}
\label{sec:eval_details}

For evaluation dataset, we use the following datasets
\begin{itemize}[left=2pt]
    \item \textbf{General Knowledge \& Instruction Following}: 
    % ARC-C ~\citep{allenai:arc}, 
    MMLU ~\citep{hendrycks2020measuring}, 
    MMLU-Pro~\citep{wang2024mmluprorobustchallengingmultitask}, 
    % MMLU-Redux ~\citep{gema2025mmlu} , 
    GPQA-Diamond \citep{rein2023gpqagraduatelevelgoogleproofqa}, IFEval~\citep{zhou2023instructionfollowingevaluationlargelanguage}, CEval~\citep{cevalhuang2023cevalmultilevelmultidisciplinechinese}, LiveBench~\citep{white2025livebench}.
    \item \textbf{Mathematics \& Reasoning}: GSM8K~\citep{cobbe2021gsm8k} , MATH-500~\citep{hendrycks2020measuring}, AIME-24~\citep{aime24}, AIME-25~\citep{dekoninck2026matharena}, LMB-Hard~\citep{fan2024hardmathbenchmarkdatasetchallenging}, ZebraLogic~\citep{zebralogic2024}.
    \item \textbf{Code Generation}: HumanEval~\citep{humanevalchen2021codex}, MBPP~\citep{austin2021programsynthesislargelanguage}, LiveCodeBench-v6~\citep{jain2024livecodebenchholisticcontaminationfree}, Codeforces~\citep{penedo2025codeforces}.
    \item \textbf{Multilingual}: MMMLU-lite~\citep{wang2024mmluprorobustchallengingmultitask}, INCLUDE-lite~\citep{romanou2024include}, MT-AIME2024~\citep{son2025linguistic}, MLogiQA~\citep{zhang2025p}.
\end{itemize}

For evaluation, we apply the official Qwen3 chat template~\citep{yang2025qwen3}. For mathematics benchmarks, we append the instruction \texttt{``Please reason step by step, and put your final answer within \textbackslash boxed\{\}.''} to each prompt. For multiple-choice benchmarks, we use \texttt{``Please show your choice in the answer field with only the choice letter, e.g., \textbackslash"answer\textbackslash": \textbackslash"C\textbackslash".''}

\subsection{Hyperparameters}
\label{app:hyperparameter}

\begin{table}[ht]
\caption{Hyperparameter settings for General-Purpose OPDLM training.}
\label{app:tab:hyperparameter_general}
\centering
\begin{tabular}{l c}
    \toprule
    \textbf{Hyperparameter} & \textbf{Value} \\
    \midrule
    Number of data epochs                        & 1 \\
    Tasks per rollout                            & 128 \\
    Effective training batch size                & 8 \\
    Learning rate                                & $1 \times 10^{-5}$ \\
    Final learning rate                          & $1 \times 10^{-6}$ \\
    Learning rate schedule                       & Cosine with warmup \\
    Warmup steps                                 & 5 \\
    \midrule
    Rollout block size                           & 4 \\
    Rollout temperature                          & 1.0 \\
    Initial maximum rollout length                    & 100 \\
    Final maximum rollout length                      & 4{,}000 \\
    Rollouts to reach final generation length    & 10 \\
    Remasking strategy                           & Dynamic \\
    Dynamic remasking confidence threshold       & 0.9 \\
    \midrule
    Divergence                                   & Forward KL \\
    \bottomrule
\end{tabular}
\end{table}

\begin{table}[ht]
\caption{Hyperparameter overrides for OPDLM-MATH training. All other hyperparameters follow \cref{app:tab:hyperparameter_general}.}
\label{app:tab:hyperparameter_math}
\centering
\begin{tabular}{l c}
    \toprule
    \textbf{Hyperparameter} & \textbf{Value} \\
    \midrule
    Number of data epochs                        & 10 \\
    Final maximum rollout length (non-thinking/thinking)                 & 2{,}000/8{,}000 \\
    Rollouts to reach final generation length (non-thinking/thinking)    & 10/20 \\
    \bottomrule
\end{tabular}
\end{table}

\cref{app:tab:hyperparameter_general} lists the hyperparameters for general-purpose OPDLM. For OPDLM-MATH (\cref{sec:posttrain}), we override only the parameters in \cref{app:tab:hyperparameter_math}; non-thinking and thinking variants differ only in rollout length.

\subsection{FLOPs Calculation}
\label{sec:flop_calc}

We follow the following FLOP estimation~\citep{kaplan2020scaling}: a forward pass through a model with $N$ parameters costs approximately $2N$ FLOPs per token, and a forward + backward pass costs approximately $6N$ FLOPs per token. In OPDLM, the teacher is queried in inference mode (forward only) and the student is updated via gradient descent (forward + backward), so the total training FLOPs are estimated as

\begin{equation}
\text{FLOPs} \;\approx\; 2 N_{\text{teacher}} \cdot T \;+\; 6 N_{\text{student}} \cdot T,
\end{equation}
where $T$ is the total number of training tokens, computed as the number of training samples times their generation length. Generation lengths follow a curriculum schedule (\cref{app:tab:hyperparameter_general} and~\cref{app:tab:hyperparameter_math}), and $T$ is summed over all rollout stages. We note that the student forward pass during training operates in block-diffusion mode at roughly $1.5$ tokens/step rather than $1$ token/step; the formula above conservatively assumes one student forward per token, slightly overestimating the FLOPs charged to OPDLM.

\subsection{Compute Usage}
\label{sec:compute_usage}
Compute Usage. Final model checkpoints were trained on H200 GPUs, while ablations and intermediate experiments were conducted on A6000 and A100 GPUs. Notably, our training pipeline is lightweight enough that all models below 4B parameters can be trained on as few as 2 A6000 GPUs.

\section{Limitations, Future Directions and Broader Impacts}
\label{sec:limitations}

\paragraph{Limitations}
Although \methodname{} achieves competitive performance on a wide range of benchmarks, there is still room for improvement on certain tasks compared to the strongest baselines. 
We believe that improving data quality is a key direction for further performance gains.
However, the pretraining datasets used by most baselines are not publicly available, making controlled comparisons difficult. 
Furthermore, while we explore thinking-mode distillation in the specialized math setting, expanding this reasoning distillation to broader, more general datasets remains an area for future work.

% Data composition limits code performance.
% Thinking distillation only in specialized setting

\paragraph{Future Directions}
\begin{itemize}[left=2pt]
    \item \textbf{Data preparation.} Our experiments use a $\sim$60K-prompt corpus assembled from publicly available sources. Building a high quality and more diverse training corpora is a promising direction. Given OPDLM's strong data efficiency, investments in data curation are likely to introduce substantial performance gains.
    \item \textbf{Reasoning distillation at scale.} We explored thinking-mode distillation only in the specialized math setting (\cref{sec:posttrain}), where it substantially boosts performance on hard reasoning benchmarks (e.g., AIME24 reaching 50\% at 8B). Extending chain-of-thought distillation to general-domain training would test whether OPDLM can transfer richer reasoning behaviors beyond mathematics, and is a natural next step toward a fully general thinking-enabled diffusion LLM.
    \item \textbf{Cross-size and cross-family distillation.} All main experiments use self-distillation between same-size, same-family teacher and student models, which may limit quality of the supervision signal. A preliminary ablation (\cref{app:teacher_size}) suggests that naively switching to larger teachers may not automatically help, motivating a more careful study of how teacher--student model size mismatches affect the final performances. Distillation across model families (e.g., Llama and Qwen models) is also worth investigating.

\end{itemize}

\paragraph{Broader Impact} Our work studies the efficient conversion of autoregressive language models (ARLMs) into diffusion language models (DLMs), and carries implications across both efficiency and societal dimensions. First, OPDLM substantially reduces the training cost of obtaining a DLM, requiring up to $7000\times$ fewer training tokens than from-scratch DLM pre-training. This directly lowers the energy footprint of producing competitive generative models, contributing toward more sustainable development of large-scale AI systems. Second, ARLMs decode one token per step, which becomes prohibitive for long-form generation. By framing ARLM-to-DLM conversion as a post-training procedure, OPDLM offers a natural path to faster inference through parallel multi-token decoding, further reducing the energy cost of deployment at scale. Finally, OPDLMs inherit the broader risks associated with generative AI systems, including the potential to reflect or amplify biases present in pre-training and post-training data. We encourage practitioners deploying OPDLMs to apply standard mitigations such as bias auditing, content filtering, and downstream alignment.

% \section{LLM Usage Declaration}
% \label{sec:llm_usage}
% \paragraph{LLM Usage.} Large language models were used solely to assist with writing and refinement of the manuscript. They were not used for research ideation, methodology, experiments, or analysis of results.

%%%%%%%%%%%%%%%%%%%%%%%%%%%%%%%%%%%%%%%%%%%%%%%%%%%%%%%%%%%%

\newpage

\end{document}